\documentclass[lettersize,journal]{IEEEtran}
\usepackage{amsmath,amsfonts}
\usepackage{algorithmic}
\usepackage{algorithm}
\usepackage{array}
\usepackage[caption=false,font=normalsize,labelfont=sf,textfont=sf]{subfig}
\usepackage{textcomp}
\usepackage{stfloats}
\usepackage{url}
\usepackage{verbatim}
\usepackage{graphicx}
\usepackage{cite}
\usepackage{booktabs}
\usepackage{makecell}
\usepackage{colortbl}
\usepackage{color}
\usepackage{threeparttable}
\usepackage{amssymb}
\usepackage{hyperref}
\usepackage{multirow}

\definecolor{mygray}{gray}{.9}
\definecolor{mypink}{rgb}{.99,.91,.95}
\definecolor{mycyan}{cmyk}{.3,0,0,0}
\definecolor{mylime}{rgb}{.75, 1, 0}

\begin{document}

\title{DREAM: A Benchmark Study for Deepfake photoREalism AssessMent}

\author{
Bo Peng, ~\IEEEmembership{Member,~IEEE,}
Zichuan Wang,
Sheng Yu,
Xiaochuan Jin,
Wei Wang, ~\IEEEmembership{Member,~IEEE,}
Jing Dong, ~\IEEEmembership{Senior Member,~IEEE,}
\thanks{The authors are with the New Laboratory of Pattern Recognition~(NLPR), Institute of Automation, Chinese Academy of Sciences~(CASIA), Beijing 100190, China, and also with the School of Artificial Intelligence, University of Chinese Academy of Sciences, Beijing 100049, China (E-mail: \{bo.peng, wwang, jdong\}@nlpr.ia.ac.cn. Jing Dong is the corresponding author.} 
\thanks{This work is supported by the National Key Research and Development Program of China under Grant No. 2024YFF0907202, National Natural Science Foundation of China under No. 62272460, 62372452, the Joint Research Project on Integration of Culture and Science and Technology between Chinese Academy of Sciences and Hunan Province \#2024JK4001, and Special Fund for Key Program of Science and Technology of Jiangsu Province under No. BG2024042.
}
}

\markboth{Journal of \LaTeX\ Class Files,~Vol.~14, No.~8, August~2021}%
{Shell \MakeLowercase{\textit{et al.}}: A Sample Article Using IEEEtran.cls for IEEE Journals}


\maketitle

\begin{abstract}
Deep learning based face-swap videos, widely known as deepfakes, have drawn wide attention due to their threat to information credibility. Recent works mainly focus on the problem of deepfake  detection that aims to reliably tell deepfakes apart from real ones, in an objective way. On the other hand,  the subjective perception of deepfakes, especially its computational modeling and imitation,  is also a significant problem but lacks adequate study. In this paper, we focus on the photorealism assessment of deepfakes, which is defined as the automatic assessment of deepfake photorealism that approximates human perception of deepfakes. It is important for evaluating the quality and deceptiveness of deepfakes which can be used for predicting the influence of deepfakes on Internet, and it also has potentials in improving the deepfake generation process by serving as a critic. This paper promotes this new direction by presenting a comprehensive benchmark called DREAM, which stands for Deepfake photoREalism AssessMent. It is comprised of a deepfake video dataset of diverse quality, a large scale annotation that includes 140,000 photorealism scores and textual descriptions obtained from 3,500 human annotators, and a comprehensive evaluation and analysis of 18 representative photorealism assessment methods, including recent large vision language model based methods and a newly proposed description-aligned CLIP method. The benchmark and insights included in this study can lay the foundation for future research in this direction and other related areas. We make the dataset available to the research community at \href{https://github.com/bomb2peng/DREAM-A-Benchmark-Study-for-Deepfake-photoREalism-AssessMent}{https://github.com/bomb2peng/DREAM-A-Benchmark-Study-for-Deepfake-photoREalism-AssessMent}.
\end{abstract}


\section{Introduction}
The emergence of Deepfake began in 2017, when a Reddit user with the name ``deepfakes" started sharing face-swapped pornography videos and movie clips, 
and it immediately drew widespread attention due to its potential harmful use against information security and personal reputation. The term deepfake later has expanded meanings that also include fully-generated human facial images, talking face videos, and synthetic audios etc.. To battle deepfakes, image forensics researchers have proposed various detection methods \cite{Seow2022AOpportunities, Mirsky2021TheDeepfakes} that classify a questioned video or image into real or deepfake, and large improvements have been made in this area. 
%
\begin{figure}[ht]
\centering
\centerline{\includegraphics[width=0.5\textwidth]{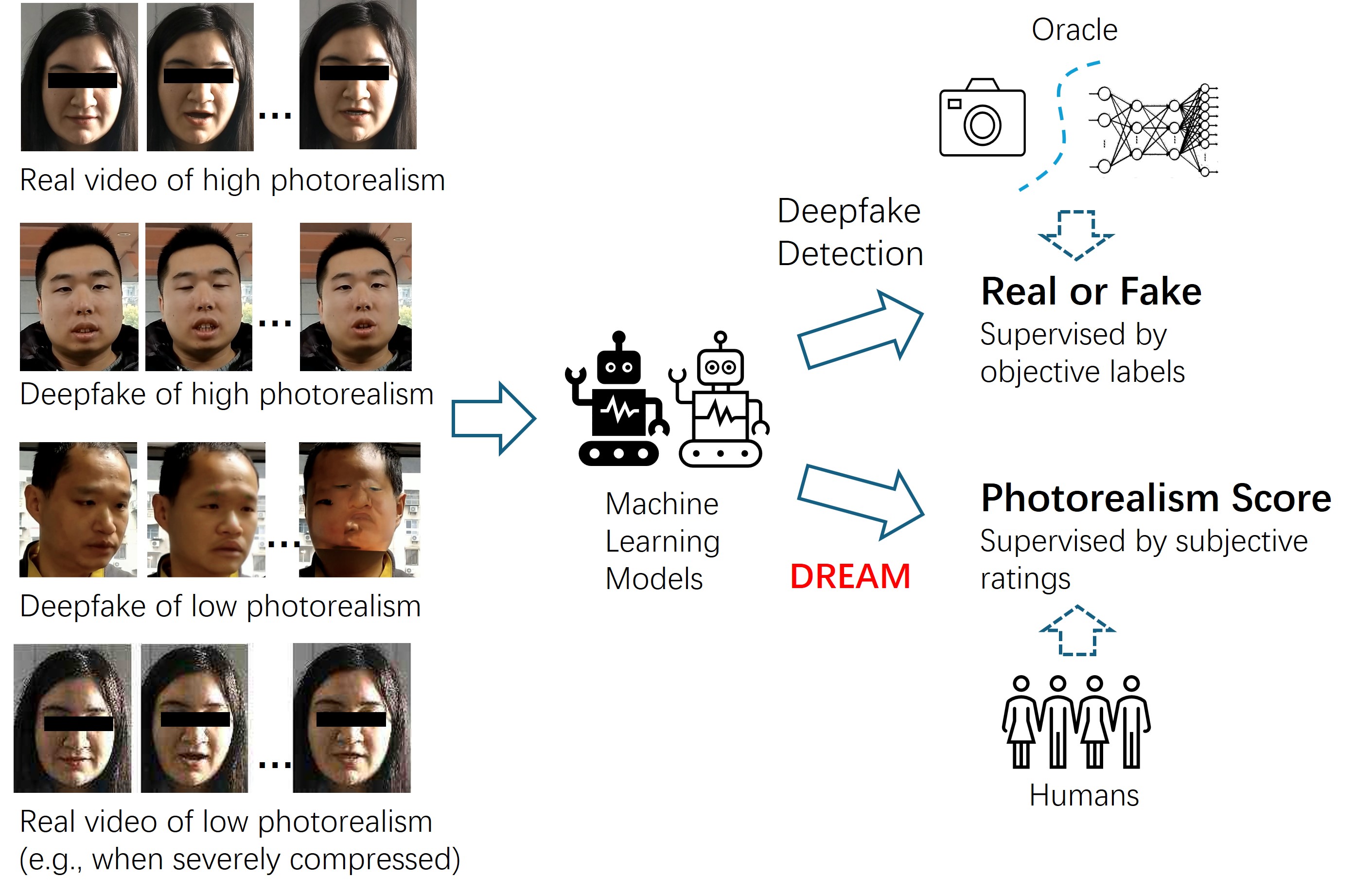}}
\caption{The difference between the traditional Deepfake detection task and the DREAM task.}
\label{fig_task}
\end{figure}

In this paper, we focus on a new task named as Deepfake photoREalism AssessMent, or DREAM for short. The difference between DREAM and the traditional deepfake detection task is illustrated in Fig. \ref{fig_task}. They both train machine learning models to predict some labels or scores from input videos, but the deepfake detection model outputs the probability of the video being a deepfake, whereas the DREAM model outputs the score of photorealism. The training of the deepfake detection model requires objective labels as ``real" and ``fake", which are provided by an oracle who knows the accurate source of each video (oftentimes the dataset creator). On the contrary, the training of the DREAM model requires subjective labels like ``very high sense of photorealism", ``average sense of photorealism", ``relatively low sense of photorealism", etc., which are provided by human raters, and the scores can be averaged over a crowd to reflect the average photorealism perception, i.e. the Mean Opinion Score (MOS). The usage of these models is also different in that, the deepfake detection models help us to judge the realness of a video, while the DREAM models imitate human perception to assess the photorealism of a video automatically. A facial video can have very high photorealism while being a deepfake in the same time. On the other hand, a real video may also be misinterpreted as being a deepfake or having low photorealism if it has very low video quality, e.g. when the video is severely compressed.
The DREAM models have potential applications in assessing the quality and deceptiveness of deepfakes as an important evaluation metric, and also have potentials in improving deepfake photorealism as a GAN-style critic, though these applications are not in the scope of this work.

In the scope of deepfake photorealism assessment, Sun et al. \cite{Sun2023VisualVideos} first attempted to use machine learning algorithms to regress human rated photorealism scores, and Peng et al. \cite{Peng2023DFGC-VRA:Assessment} promoted this new task by organizing a deepfake photorealism assessment competition. However, these two previous works are conducted on a dataset that has very limited annotation, since each video is only rated by 5 human viewers. This is far from enough in comparison with the related field of natural video quality assessment that typically has tens to hundreds of ratings for each video \cite{Tu2021UGC-VQA:Content}. Apart from the insufficient annotation problem, the previous works also lack comprehensive comparison and in-depth analysis of DREAM methods, especially when considering the current trend of large model based methods and multi-modal understanding.

This paper is an extension of our previous work \cite{Peng2023DFGC-VRA:Assessment, Sun2023VisualVideos}. It presents a comprehensive DREAM benchmark, comprising a deepfake video dataset of diverse quality with large scale annotations of photorealism scores and textual descriptions, a thorough analysis of the collected annotations, and comprehensive experiments and analyses of extended representative  methods that also include recent ones based on Vision-Language Models (VLM). Specific improvements of this paper over our previous work \cite{Peng2023DFGC-VRA:Assessment, Sun2023VisualVideos} are summarized in two aspects: 
\begin{enumerate}
    \item Improvement and extension of the annotation protocol for the DREAM dataset, which includes a great increase in the number of annotators, additional annotation of textual descriptions, and more in-depth dataset analysis.
    \item We propose a new method DA-CLIP, profiting from the new textual descriptions to learn a description-aligned representation space, and it achieves the best in-domain performance and good explainability.
\end{enumerate}

\section{Related Work}
Our work is related to several area of existing work, and they are compared in high-level in Table \ref{tab_related}. We elaborate on these methods in the following.
\begin{table}[htb]
    \centering
    \caption{\centering High-level comparison of related work.}
    \scriptsize
    \begin{tabular}{c c c}
    \hline
        \textbf{Methods} & \textbf{Advantages} & \textbf{Dis-advantages} \\ 
        \hline
        \makecell{Deepfake \\Detection}   & \makecell{Distinguish deepfake from \\an objective perspective}    & \makecell{Lack explainability or\\ explanations are inaccurate} \\
        \hline
        \makecell{Handcrafted \\IQA/VQA}    & \makecell{Suitable for small datasets and \\simple degradation types}  & \makecell{Handcrafted features lack \\learning ability}\\
        \hline
        \makecell{Deeplearning \\based \\IQA/VQA}   & \makecell{Suitable for large datasets, \\and end-to-end feature learning}  & \makecell{Efficiency problem on \\high-res videos} \\
        \hline
        \makecell{VLM \\based \\IQA/VQA}   & \makecell{Leveraging rich knowledge in \\large vision-language models}     & \makecell{Based on predefined sentence \\templates, cannot make use of \\detailed textual descriptions}\\
        \hline
        FIQA   & \makecell{For facial image usability \\to facial recognition systems}  & \makecell{Cannot apply to other domains \\outside face images}\\
        \hline
        AGIQA   & \makecell{Evaluate the quality of AI-\\generated images}  & \makecell{No work on facial \\deepfake video assessment} \\
        \hline
        Our work   & \makecell{Tackle  photorealism assessment\\ of deepfake videos, with \\good explanation ability}  & \makecell{Did not consider domains \\outside facial videos and \\no audio modality} \\
        \hline
    \end{tabular}
    \label{tab_related}
\end{table}
\subsection{Deepfake Detection}
Deepfake detection aims at distinguishing whether a
facial image/video is deepfake or real. With the availability
of recent benchmarks and large datasets \cite{Khalid2021FakeAVCeleb:Dataset, Zhou2021FaceWild, Le2021Openforensics:In-the-wild},
deepfake detection models have obtained better performance,
by employing self-supervised data augmentations
\cite{Shiohara2022DetectingImages}, stronger models like Transformers \cite{Zhao2023ISTVT:Detection}, and audio-visual consistency modeling
\cite{Yang2023AVoiD-DF:Deepfake} etc.. However, they still struggle in generalizing to detecting unseen deepfake methods, and the lack of explainability also hinders their real-world usage \cite{Peng2022CounterfactualDeepfakes} in law enforcement or court of law.

To tackle these problems, notable recent progress includes the employment of Vision-Language Models (VLM) \cite{Foteinopoulou2024AReasoning, Zhang2024CommonDetection, Yu2025UnlockingDetection, Sun2025TowardsDetection} to introduce textual explanations besides the common real or fake labels. In the work of \cite{Foteinopoulou2024AReasoning}, an evaluation of off-the-shelf VLMs was conducted to test their abilities in deepfake detection and more fine-grained tasks like multi-choice and open-ended visual question answering. In \cite{Zhang2024CommonDetection}, the authors annotated the FaceForensics++ \cite{Rossler2019Faceforensics++:Images} dataset with human-identifiable fake features as textual explanations and propose to train a VLM for the Deepfake Detection Visual Question Answering (DD-VQA) task. The work \cite{Yu2025UnlockingDetection} adopted similar VLM based question answering methodology, but their ground-truth textual annotations are automatically obtained with simulated self-blended \cite{Shiohara2022DetectingImages} face forgery images and mainly describe the forgery regions. In \cite{Sun2025TowardsDetection}, a CLIP \cite{Radford2021LearningSupervision} based multi-modal contrastive learning method was proposed, and fine-grained textual annotations describing forgery types are obtained by detecting several pre-defined common traces, apart from those describing forgery regions as in \cite{Yu2025UnlockingDetection}.
In these VLM based deepfake detection works, although textual descriptions of forgery regions and types are output to augment explainability, their ultimate goal is still the objective classification of real and fake. On the other hand, the DREAM task in this paper focuses on the subjective photorealism rating. 

There are also some papers \cite{Korshunov2021SubjectiveVideos, Lu2023SeeingImages} discussing the discrepancies between detection models and human perception of visual deepfakes. They found the deepfake traces may oftentimes be not perceivable by humans and still be classified as fake by models, while there are also perceptually obvious fake samples that can escape detection models.
Apart from visual deepfake, analysis on human perception of audio deepfake and its machine learning detectors is conducted in \cite{Muller2022HumanDeepfakes, Warren2024BetterDetectors}, and the perception of cross-modal audiovisual deepfake is studied in \cite{Hashmi2024UnderstandingInsights}. We note that the perception of audio and audiovisual deepfakes is broadly related to our work, but we focus on the visual deepfake in this work.
\subsection{Image and Video Quality Assessment} \label{sec-Related-IQA}
Image and video quality assessment, i.e., IQA and VQA,  are classical research topics in image processing and multimedia community. They primarily aim at assessing the subjective visual quality of natural images and videos when they go through some degradation processes, e.g., lossy compressions
and network streaming, or when they are captured in bad conditions. IQA and VQA methods can be classified into full-reference, reduced-reference, and no-reference methods, depending on the availability of original images or videos as references. We only introduce some no-reference IQA/VQA methods here as they are the most related. 

\textbf{Handcrafted IQA} features like BRISQUE \cite{Mittal2012No-referenceDomain}, GM-LOG \cite{Xue2014BlindFeatures}, and HIGRADE \cite{Kundu2017No-referencePictures} extract image quality related features based on the Natural Scene Statistics (NSS) model, where they use different filters on the image and extract designed statistics as features. IQA features are useful for assessing image-level degradations such as noise, blurring, and compression artifact, but cannot represent motion artifacts in videos like jerkiness. This calls for handcrafted VQA features, e.g., TLVQM \cite{Korhonen2019Two-levelAssessment} and V-BLIINDS\cite{Saad2014BlindQuality}. These methods include features extracted from motion vectors between two consecutive frames or from frame differences using different design. RAPIQUE \cite{Tu2021RAPIQUE:Content} further combines hand-crafted spatial and temporal features with pretrained CNN feature followed by a regression head for better performance. VIDEVAL \cite{Tu2021UGC-VQA:Content} also proposes to fuse and select effective features from different NSS models to balance VQA performance and efficiency.

\textbf{Deep-learning based IQA/VQA} methods become more popular with the end-to-end feature learning ability. To tackle the problem of lacking large annotated training datasets, RankIQA \cite{Liu2017Rankiqa:Assessment} proposes to train on pairs of images with Siamese network and ranking loss. Patch-VQ \cite{Ying2021Patch-VQ:Problem} proposes the space-time video quality mapping network based on video patch (`v-patch') level annotations.
To reduce the computational cost that hinders end-to-end VQA model training, FastVQA \cite{Wu2022FAST-VQA:Sampling} proposes Grid Mini-patch Sampling (GMS) and Fragment Attention Network (FANet). 

\textbf{VLM based IQA/VQA} methods are introduced recently, including CLIP-IQA \cite{Wang2023ExploringImages}, Q-Align \cite{Wu2024Q-ALIGN:Levels} and 
DeQA-Score \cite{You2025TeachingDistribution} that benefit from the strong capability and prior knowledge in VLMs. 
The CLIP-IQA method \cite{Wang2023ExploringImages} is based on an off-the-shelf Contrastive Language-Image Pretraining (CLIP) \cite{Radford2021LearningSupervision} model and does not require task-specific finetuning, making it a very flexible and versatile method. It employs antonym prompt pairs to obtain a binary classification output which serves as zero-shot assessment of the look and feel of images in many aspects. Since CLIP-IQA is training-free, its performance may fall short. Then, the CLIP-IQA+ is an extension method by introducing Context Optimization (CoOp) \cite{Zhou2022LearningModels} to finetune the input prompts on the training set and can obtain better results. 
CLIP-IQA has good flexibility but lacks in performance, and Q-Align \cite{Wu2024Q-ALIGN:Levels} is designed to learn the discrete quality levels denoted by texts using a VLM to strive for better performance. It requires finetuning, during which the loss function is just the softmax loss for predicting the next token from large language model. However, Q-Align still has the problem of discretization error when converting the continuous MOS to the nearest integer as groundtruth. To solve this, instead of using a single hard label, DeQA-Score \cite{You2025TeachingDistribution} uses soft labels obtained from fitting a Gaussian distribution for the quality score, with MOS as the mean and the standard deviation of opinion scores as the std. Then, the KL-divergence loss is used to finetune the VLM in replacement of the softmax loss of Q-Align. 
A problem of existing VLM based methods is their input and output prompts are based on predefined sentence templates, and they cannot make full use of detailed textual description that is available in our dataset.

\textbf{Face Image Quality Assessment (FIQA)} is another related but different area  \cite{Schlett2022FaceSurvey, Boutros2023CR-FIQA:Classifiability}, which specifically focuses on the utility of face images for downstream facial recognition (FR) applications instead of the perceptual visual quality. Apart from blur, noise and illumination that are common affecting factors in traditional IQA, FIQA also considers the influence of facial pose, expression and occlusion \cite{Schlett2022FaceSurvey}. FIQA methods include regression based ones and unsupervised ones rooted in FR model responses \cite{Boutros2023CR-FIQA:Classifiability}.
%
\subsection{AI-Generated Image Quality Assessment (AGIQA)}
Traditional IQA/VQA works mainly focus on natural scene images and videos, and there is a new trend in the quality assessment of AI generated imagery (or AGIQA), e.g., GAN and Diffusion generated images. These generated images are commonly evaluated using the Fréchet Inception Distance (FID) metric and alike ones, which measures the distance between real and fake image feature distributions. However, FID cannot indicate the visual quality of each individual image. GIQA \cite{Gu2020Giqa:Assessment} addresses this problem by proposing several models for predicting the quality of individual GAN images, with the best model being a Gaussian Mixture Model. The work \cite{Tian2022GeneralizedImagesb} proposes generalized visual quality assessment for face images generated by various GANs, employing meta-learning and pair-wise ranking on pseudo quality scores to mitigate overfitting. 

Recent works tackle this problem by proposing more large-scale datasets with human annotated MOS scores or preference ranking, including AGIQA-3k \cite{Li2024AGIQA-3K:Assessment}, PKU-I2IQA \cite{Yuan2023PKU-I2IQA:Images}, ImageReward \cite{Xu2023ImageReward:Generation}, etc.. The quality evaluation dimensions may consider photorealism, quality, local defects, text-image alignment, aesthetic, and even harmlessness. The assessment methods are similar to those for natural images and videos, and some also adopt VLM based methods \cite{Xu2023ImageReward:Generation, You2025TeachingDistribution}. The main difference of generated image quality assessment from the natural counterpart is that it has to additionally consider the image and prompt alignment. Besides, the quality-impacting factors in the visual aspect are also different, where generated images have more structural and textural defects resulting from the generation process that are absent in the natural images. 
Meanwhile, AI generated video quality assessment starts to emerge \cite{Zhang2024BenchmarkingModel}, but it is relatively under-studied compared to generated image assessment, because general domain video generation still has large quality gap from real ones. On the other hand, deepfake videos, especially face-swap videos, have achieved deceiving high qualities in their best form, but this area still lacks targeted quality assessment studies, especially in the photorealism aspects.
%
\section{Improvement and extension of annotation protocol}
\subsection{The Annotation Process}
\begin{figure*}[ht]
\centering
\centerline{\includegraphics[width=1.0\textwidth]{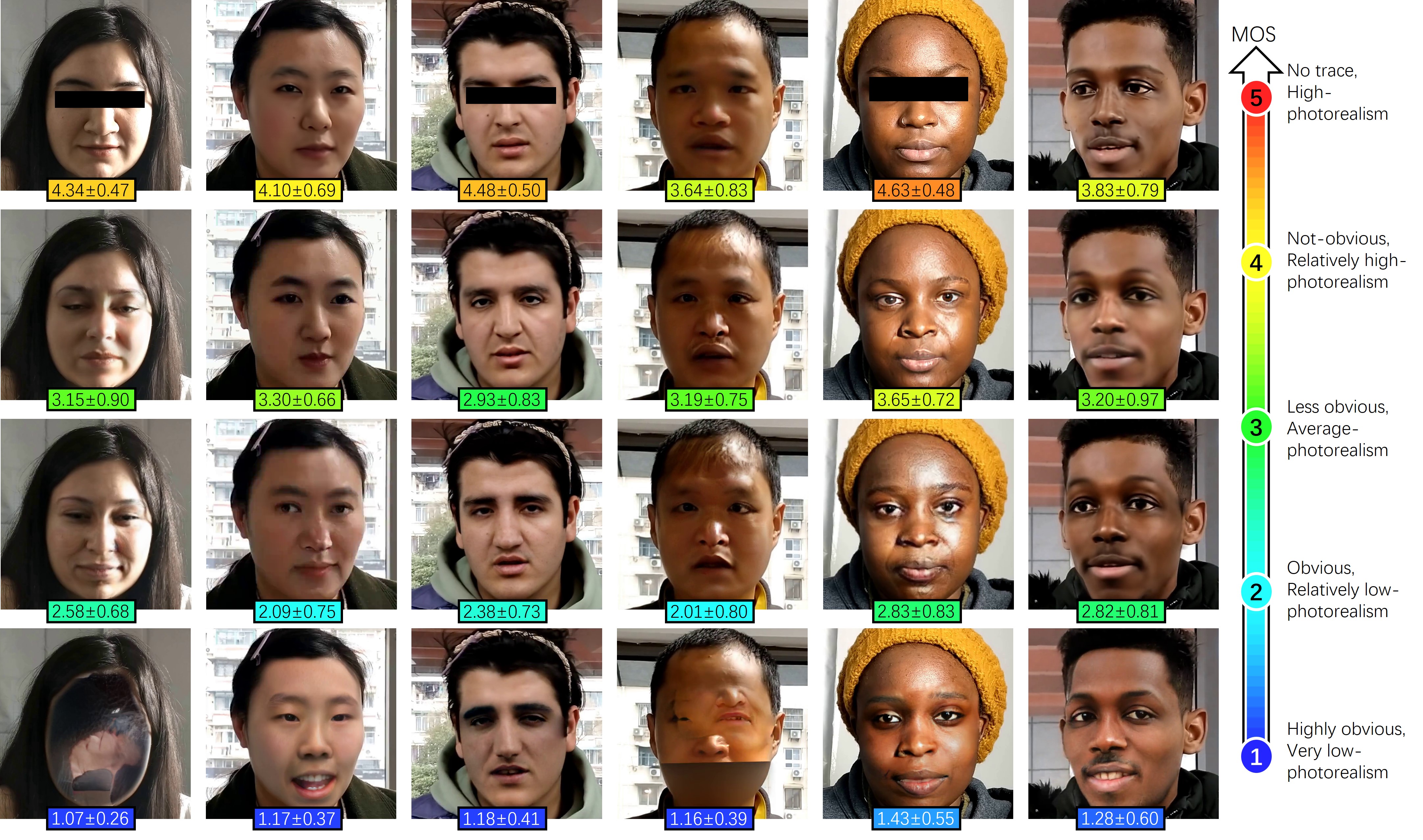}}
\caption{Dataset samples with MOS$\pm$std of annotated photorealism scores. Note the annotations are based on videos, whereas we can only show video frames here. The last row of the 1st, 3rd, and 5th columns are obscured to protect privacy in real frames, and the rest are all deepfake frames. Enlarge in the digital version for better view.}
\label{fig_samples}
\end{figure*}
The deepfake dataset we annotate is from the DFGC-2022 \cite{Peng2022DFGCCompetition} dataset, which was created using various face-swap methods and has diversified degrees of photorealism, as shown in Fig. \ref{fig_samples}. More specifically, it contains face-swap videos for 20 pairs of people with balanced genders and skin-colors. The total number of deepfake creation methods in this dataset is 35, which includes popular deepfake tools like DeepFaceLab \cite{GitHubDeepfakes.}, FaceShifter \cite{Li2019FaceShifter:Swapping}, SimSwap \cite{Chen2020SimSwap:Swapping} etc., and variants of them with enhanced post-processing. Each video is about 5 seconds and has $1920 \times 1080$ resolution. We adopt the five-grades annotation protocol shown in Table \ref{tab_points}. 

\begin{table}[htb]
    \centering
    \caption{\centering The five-grades annotation protocol used in our dataset.}
    \scriptsize
    \begin{tabular}{ccc}
    \hline
        \textbf{Points} & \textbf{Forgery traces} & \textbf{Description} \\ 
        \hline
        1   & Highly obvious    & \makecell{Very low sense of photorealism: highly obvious \\traces can be seen, seriously affecting viewing.}\\
        \hline
        2    & Obvious  & \makecell{Relatively low sense of photorealism: obvious traces \\can be seen, which hinders normal viewing.}\\
        \hline
        3   & Less obvious  & \makecell{Average sense of photorealism: less obvious traces \\can be seen, with little impact on normal viewing.} \\
        \hline
        4   & \makecell{Not obvious \\but visible}     & \makecell{Relatively high sense of photorealism: traces are not \\obvious or are uncertain.}\\
        \hline
        5   & No trace  & \makecell{High sense of photorealism: no traces can be seen.}\\
        \hline
    \end{tabular}
    \label{tab_points}
\end{table}

Before each annotation session begins, the annotators were given a quick lecture on what is deepfake, the introduction of all photorealism grades, demo videos in each photorealism grade (with descriptions of reasons), and how the annotation process goes. The annotation system is a webpage based platform.
The annotators can view, review, pause, and put to full-screen the to-be-annotated video as they like, then select a proper photorealism grade, and finally input a textual description of the reason for this grade. We ablated the audio signal from each video to ensure the rating is purely based on the visual quality.
About the description of reason, during the lecture, it is suggested to use the form \textit{``Someplace looks like having some-artifact of some-extent"}, such as \textit{``The whole face flickers dramatically and the mouth movement looks unnatural"}, but the form is not strictly enforced and the annotators have much freedom. If an annotator gives a 5, the description will be automatically set to \textit{``The photorealism is very high, and there are no detectable signs of forgery"}. Note this setting is done at the backstage, and the annotators are still required to input some thing (describing the videoed person) to prevent lazy annotations biasing to 5 points.
To make the dataset publicly available, we collected  consents from all annotators to use their annotations and relevant information for academic experiments and analysis.\footnote{This human subjects research is approved by the Institutional Review Board (IRB) of CASIA under issue No. IA21-2310-020301.}

%
\begin{figure}[ht]
\centering
\centerline{\includegraphics[width=0.5\textwidth]{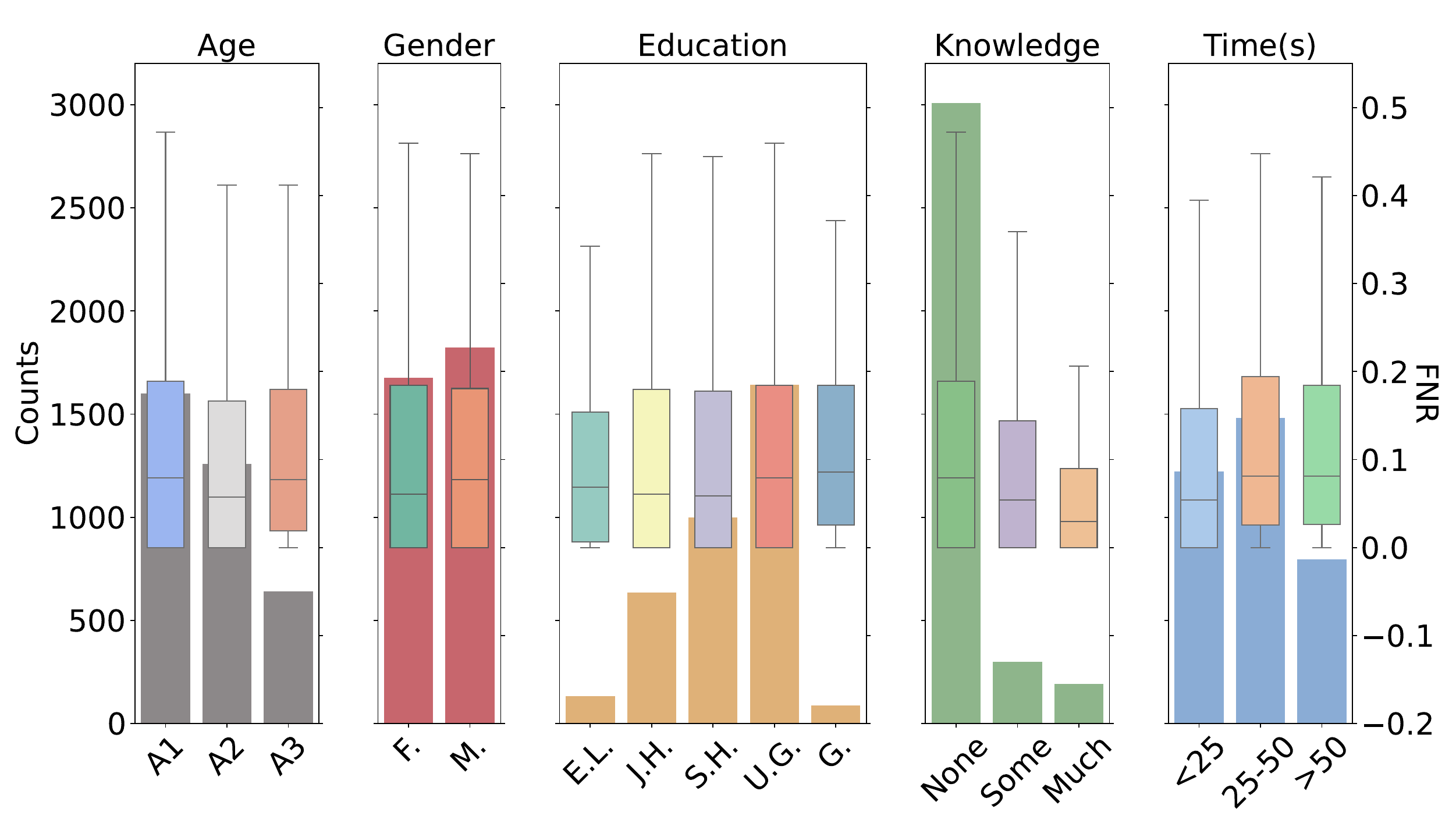}}
\caption{The distribution of annotators across age, gender, education, prior knowledge of deepfake, and average annotation time per-video. The box plot of each group's false negative rate (FNR), i.e., falsely recognizing a deepfake video as real, is also shown. A1: 18-35, A2: 36-55, A3: above 55. F.: Femal, M.: Male. E.L.: Elementary school and lower, J. H.: Junior high, S. H.: Senior high, U. G.: Undergraduate, G.: Graduate. None: Never heard of, Some: Heard related reports, Much: used similar tools.}
\label{fig_population}
\end{figure}
\begin{figure}[ht]
\centering
\centerline{\includegraphics[width=0.4\textwidth]{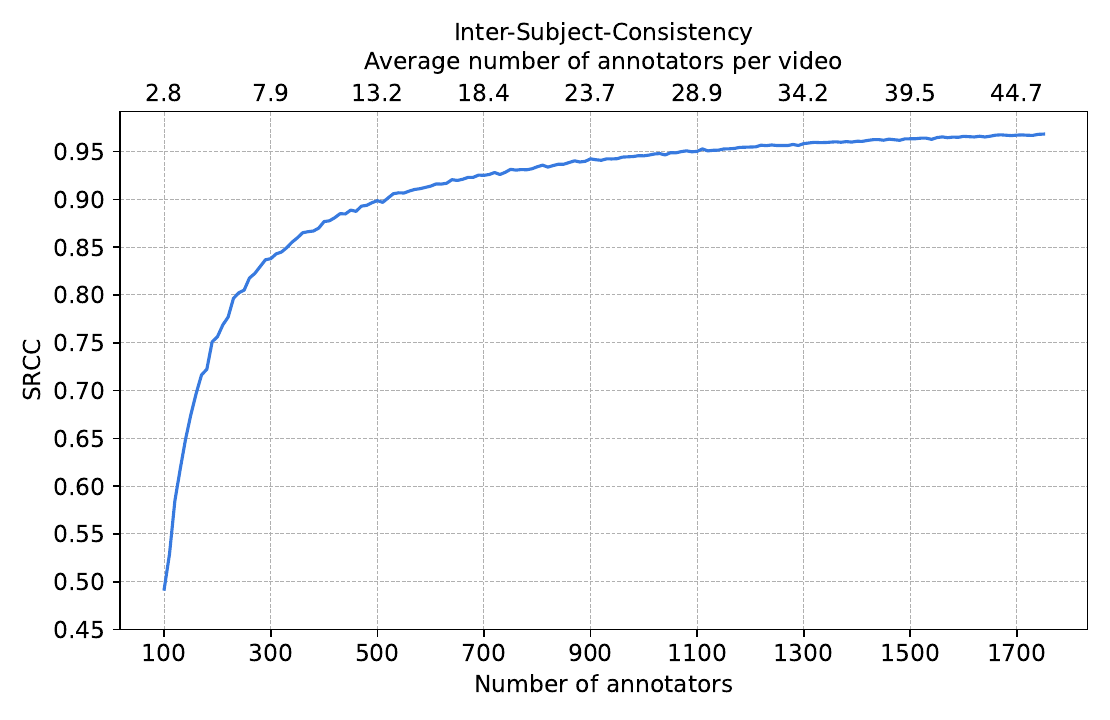}}
\caption{Mean agreement (SRCC) of MOS values under different number of annotators. When the number of annotators increases, the average number of annotations per video also increases, and the SRCC grows.}
\label{fig_inter-subjects}
\end{figure}
\begin{figure}[ht]
\centering
\centerline{\includegraphics[width=0.4\textwidth]{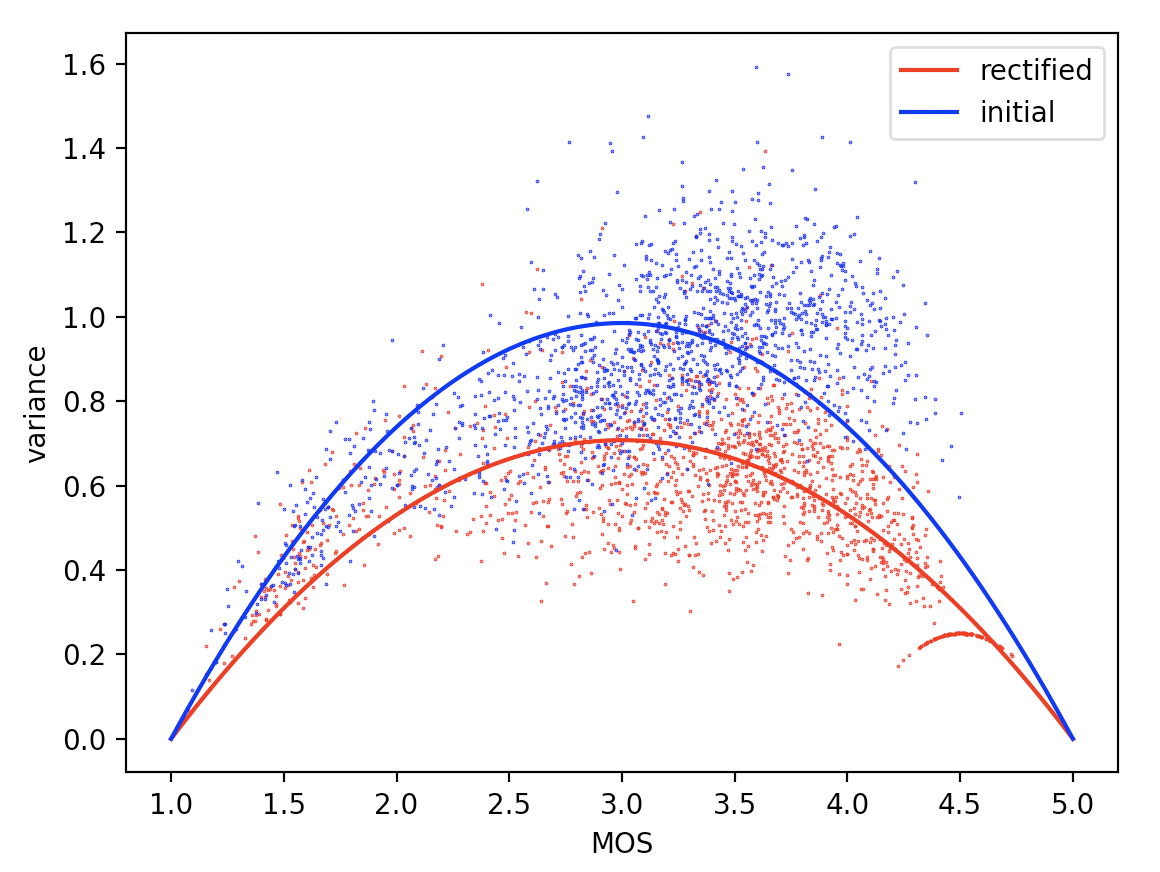}}
\caption{The variance to MOS scatter plots of all videos before and after the quality control and rectification. Each point represents a video stimulus. The quadratic regression curves are also shown.}
\label{fig_SOS}
\end{figure}

Different from our previous work \cite{Peng2023DFGC-VRA:Assessment, Sun2023VisualVideos} where only 1,400 deepfake videos from DFGC-2022 were annotated, in this work we also added 120 real videos for annotation, summing up to 1,520 videos annotated. More importantly, we greatly increased the scale of annotators to 3,500 compared to only 5 in \cite{Peng2023DFGC-VRA:Assessment, Sun2023VisualVideos}. The recruited annotators are all from China, and the distribution of relevant annotator attributes can be seen in Fig. \ref{fig_population}. As can be seen, the gender is roughly balanced, most of them are relatively young and have undergraduate education, and it should be noted that the vast majority never heard of deepfake before. In terms of annotation time on each video, most annotators can accomplish one video within 50 seconds. 

On average, each video is annotated independently by 92  annotators. The Mean Opinion Score (MOS) of each video is calculated by taking the mean of all photorealism scores this video obtains, and the MOS is used as the groundtruth in the DREAM task. To verify the scale of annotators, we examine the trend of inter-subject consistency as the number of annotators increases \cite{Hosu2020KonIQ-10k:Assessment, Ying2021Patch-VQ:Problem}, as shown in Fig. \ref{fig_inter-subjects}. At each number of annotators, we randomly sample two non-overlapping groups each of this number from the whole 3,500 annotators and calculate the MOS agreement using the Spearman Rank Correlation (SRCC) metric, and repeat this process for 10 times to obtain the mean agreement. As the number of annotators increases, the MOS agreement also increases. At the right-most point of 1,750 annotators, each video is annotated by 46.1 annotators on average, the mean SRCC over 10 times sampling is 0.9680, and the standard variation is 0.0012. This means at this scale, the MOS obtained from different groups of annotators is already very stable to serve as the groundtruth label. In our dataset, the total number of annotators is 3,500, which guarantees even better groundtruth. On the contrary, when the average number of annotations per video is only 5, as is the case in our previous work \cite{Peng2023DFGC-VRA:Assessment, Sun2023VisualVideos}, the SRCC is around 0.75, making that ``groundtruth" less credible.

\subsection{Annotation Quality Control}
To guarantee high quality of the annotation, we decreased the workload of each annotator to only 40 videos, to avoid careless mistakes from long-time tedious work. Moreover, we mixed 5 checker videos with gold standard scores into the 40 videos for checking the annotation quality. The checker videos are either real videos that should be rated 5-points or extremely low photorealism videos that should be rated 1-point. 
If an annotator makes mistake on one checker video with more than $\pm1$ point deviation, the whole annotation session will be disqualified, and this annotator will be required to take the lecture again and then re-annotate the whole session until the hidden conditions are met. Due to the task difficulty and that most people are not familiar with deepfake (see Fig. \ref{fig_population}), in total 1547 annotators went through the re-annotation process, which is 44.2\% of the total number. This also implied the necessity of our quality control step. 

To further validate the effectiveness of our quality control step, we show the annotators' score variance with respect to the MOS of all video stimuli before and after the rectification. As can be seen in Fig. \ref{fig_SOS}, the variances after rectification are clearly lower than before. We fit a quadratic regression model for the dependence of variance ($\sigma^2$) on MOS, with the form $\sigma^2({\rm MOS})=a({\rm MOS}-1)(5-{\rm MOS})$, also shown in Fig. \ref{fig_SOS}. Here, $a$ is called the SOS parameter \cite{Hosu2017TheKoNViD-1k, Hofeld2011SOS:Enough}, where SOS represents standard deviation of opinion scores. The SOS parameter quantifies the variance of annotator ratings, being respectively 0.25 and 0.18 before and after rectification, and it is the lower the better. According to \cite{Hofeld2011SOS:Enough}, normal SOS parameters for video quality assessment annotations are in the range of [0.11, 0.21], which our rectified annotation satisfied well. 

\subsection{Analysis of Gathered Annotations}
\begin{figure}[ht]
\centering
\centerline{\includegraphics[width=0.35\textwidth]{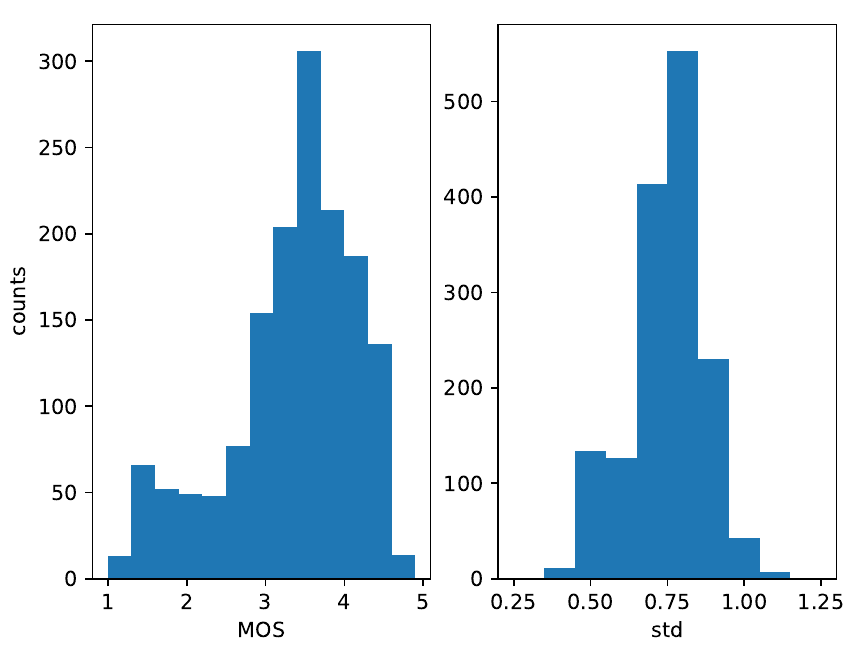}}
\caption{The histograms of MOS and standard deviation of opinion scores across videos.}
\label{fig_MOS-hist}
\end{figure}
%
\begin{figure}[ht]
\centering
\centerline{\includegraphics[width=0.5\textwidth]{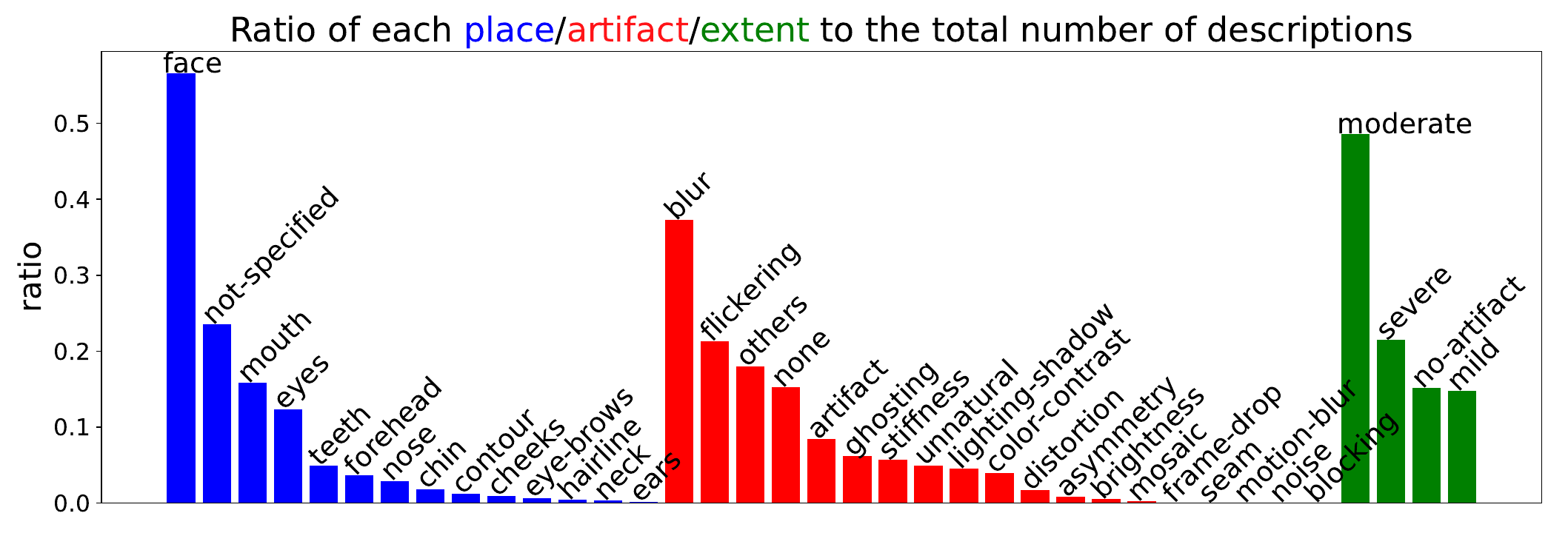}}
\caption{The distribution of described facial places, artifact types, and extents, as summarized by ChatGPT-o3 based on all annotated textual descriptions.}
\label{fig_category-hist}
\end{figure}
The distribution of the final MOS and standard deviation of opinion scores (SOS) of all annotated videos is shown in Fig. \ref{fig_MOS-hist}. We can see that most videos in this dataset have moderate to relatively high photorealism, and the standard deviation in opinion scores is relatively low with the majority under 1. Some samples of the dataset are shown in Fig. \ref{fig_samples}. 
We further employ ChatGPT-o3, a powerful OpenAI large model for reasoning, to analyze the distribution of described places, artifacts, and extents in the annotation. After careful prompting, checking, and re-prompting, we ended up with 14 categories of places, 19 categories of artifacts, and 4 categories of extents, and their ratios to the total number of descriptions in the dataset are shown in Fig. \ref{fig_category-hist}. This prompting procedure is detailed in our github repository for better transparency and reproducibility \footnote{\href{https://github.com/bomb2peng/DREAM-A-Benchmark-Study-for-Deepfake-photoREalism-AssessMent/blob/main/Prompts_categorization.md}{https://github.com/bomb2peng/DREAM-A-Benchmark-Study-for-Deepfake-photoREalism-AssessMent/blob/main/Prompts\_categorization.md}}. Note each description can include more than one kind of artifact/place/extent, thus their ratios may sum up to be over 1. We can see that most descriptions target the whole face followed by mouth and eyes. Blurring and flickering are the two most noticeable and reported artifacts, though the (descriptions of) artifact types have long tail. Finally, most annotators tend to describe the videos as having moderate extent of some artifacts.

As an interesting populational investigation, in Fig. \ref{fig_population} we also show the box plot of each group's false negative rate, i.e., falsely recognizing a deepfake video as real. Here, a deepfake video scored at 5 points is treated as a false negative case. As can be seen, a person with average perception ability can only be fooled by less than 10\% deepfake videos that are most realistic. We then run the Kruskal–Wallis statistical test on the FNRs of different groups. The p-value of the null hypothesis that the FNR median of all of the groups are equal is obtained. For age, gender, education, prior knowledge, annotation time, the p-value is respectively 0.25, 0.41, 0.64, $8.2\times10^{-8}$, $9.8\times10^{-8}$. This indicates that people with different prior knowledge of deepfake are significantly different in the ability to recognize deepfakes, and that people using different annotation time are so too. Specifically, people with
more prior knowledge of deepfake tend to be less fooled by deepfakes. It also applies to people using less annotation time, which may be because they are more confident in this task, reflecting potentially better deepfake perception ability. 

\section{Proposed Method}
\begin{figure}[th]
\centering
\centerline{\includegraphics[width=0.5\textwidth]{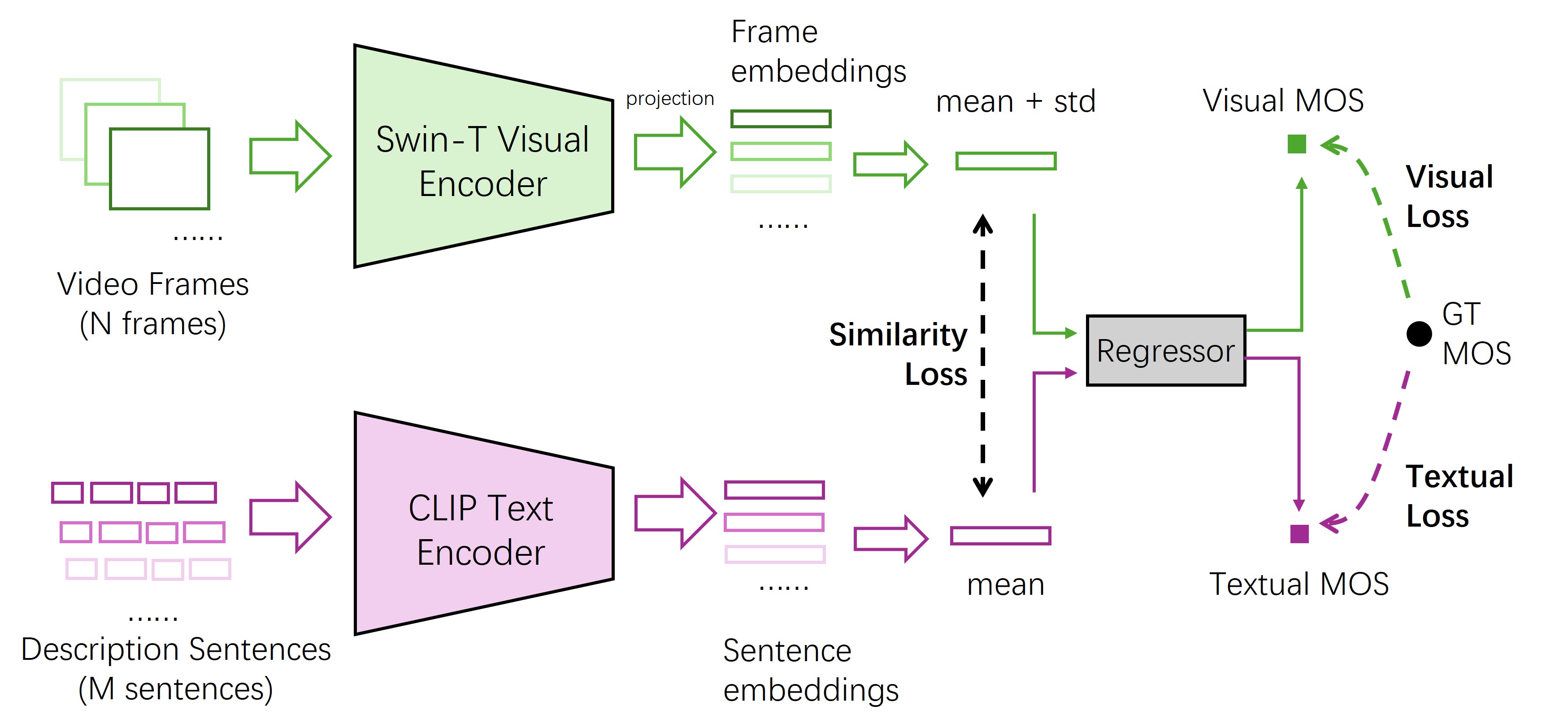}}
\caption{Illustration of the proposed DA-CLIP model. It learns a shared visual-textual representation space by aligning video features with their corresponding textual description features, and the two kinds of features are also supervised by regression losses.}
\label{fig_CLIP}
\end{figure}

We propose a new Description-Aligned CLIP Method (DA-CLIP) to achieve better performance on the DREAM task. It benefits from the cross-modal alignment ability of CLIP, as shown in Fig. \ref{fig_CLIP}.

Since the videos are annotated with textual descriptions of perceived artifacts, we can leverage this textual information to learn a shared representation between visual and textual data. We use the Swin-Transformer from the OPDAI method in \cite{Peng2023DFGC-VRA:Assessment} pretrained for deepfake detection (\texttt{swinv2\_large\_window12to16\_192to256\_22kft1k}, 197M-parameters) to obtain visual embeddings from video frames, this is because this pretrained model has proven to be very effective in the DREAM task. This visual representation is then projected to the same dimensionality as the textual representation by a fully connected projection layer. To better handle video input, we extract frame features and then the mean and std pooling is used over the $N$ frames, and they are summed to obtain the visual representation of the input video. Here, the standard deviation of frame features acts as an representation of the dynamic information in video. As for the textual stream, all sentences describing the same video are embedded by the CLIP textual encoder (OpenAI's 63M-parameter 12-layer 512-wide Transformer), and they are mean pooled over the $M$ sentences to obtain the textual representation. 

Since we aim to learn a shared visual-textual representation space, we use a unified regressor, which is a fully connected layer, to predict the MOS from either the visual or the textual representation. During training, three losses are used, i.e., the visual regression loss $L_V$, the textual regression loss $L_T$, and the cross-modal similarity loss $L_{sim}$. The two regression losses all adopt the combination of Norm-in-norm and KL-divergence loss, inspired by the OPDAI method from \cite{Peng2023DFGC-VRA:Assessment}. 
Specifically, the Norm-in-norm loss is originally proposed for image quality assessment \cite{Li2020Norm-in-normAssessment}. It uses normalization to speed-up convergence and to encourage linear predictions with respect to groundtruth scores. 
Given label $Q$ and prediction $\hat{Q}$, the Norm-in-norm loss is defined as:
\begin{align}
    & L_{NIN}(Q, \hat{Q}) = \sum_{i=1}^N \left|\hat{S}_i - S_i\right| \\
    & S_i = \frac{Q_i-\frac{1}{N}\sum_{i=1}^N Q_i}
    {(\sum_{i=1}^N \left| Q_i-\frac{1}{N}\sum_{i=1}^N Q_i \right|^q)^{\frac{1}{q}}}  \label{Eqn_S}
\end{align}
where $S_i$ is the normalized version of the groundtruth score $Q_i$, and $\hat{S}_i$ can be similarly calculated. $N$ is the number of training samples in a batch. The parameter $q$ is set to 2 here. The KL-divergence loss is defined as:
\begin{align}
    &L_{KLD}(Q,\hat{Q}) = \sum_{i=1}^N \hat{W}_i \times \log\frac{\hat{W}_i}{W_i} \\
    &W_i = \frac{\exp(Q_i)}{\sum_{i=1}^N \exp(Q_i)}  \label{Eqn_W}
\end{align}
where $W_i$ is the Softmax-normalized version of  the groundtruth scores $Q_i$, and $\hat{W}_i$ can be similarly calculated. Finally, the regression loss ($L_V$ and $L_T$) is the sum of the two losses:
\begin{equation} \label{Eqn_OPDAI}
    L(Q, \hat{Q}) = L_{NIN}(Q,\hat{Q}) + L_{KLD}(Q, \hat{Q})
\end{equation}

The cross-modal similarity loss is imposed on the visual and the textual representations to pull corresponding pairs closer, and we use the cosine similarity subtracted by 1 for this:
\begin{equation}
    L_{sim} = 1 - \sum_{i=1}^N \frac{x_i \cdot t_i}{||x_i||\cdot||t_i||},
\end{equation}
where $x_i, t_i$ is respectively the extracted visual and textual representation of a video, and $N$ is the number of training data. The total loss is:
\begin{equation}
    L_{DA} = L_V+L_T+\lambda L_{sim} .
\end{equation}
Here, $\lambda$ is set to 
0.2
for best performance, and we finetune the whole model on the training dataset. 

\begin{figure}[th]
\centering
\centerline{\includegraphics[width=0.5\textwidth]{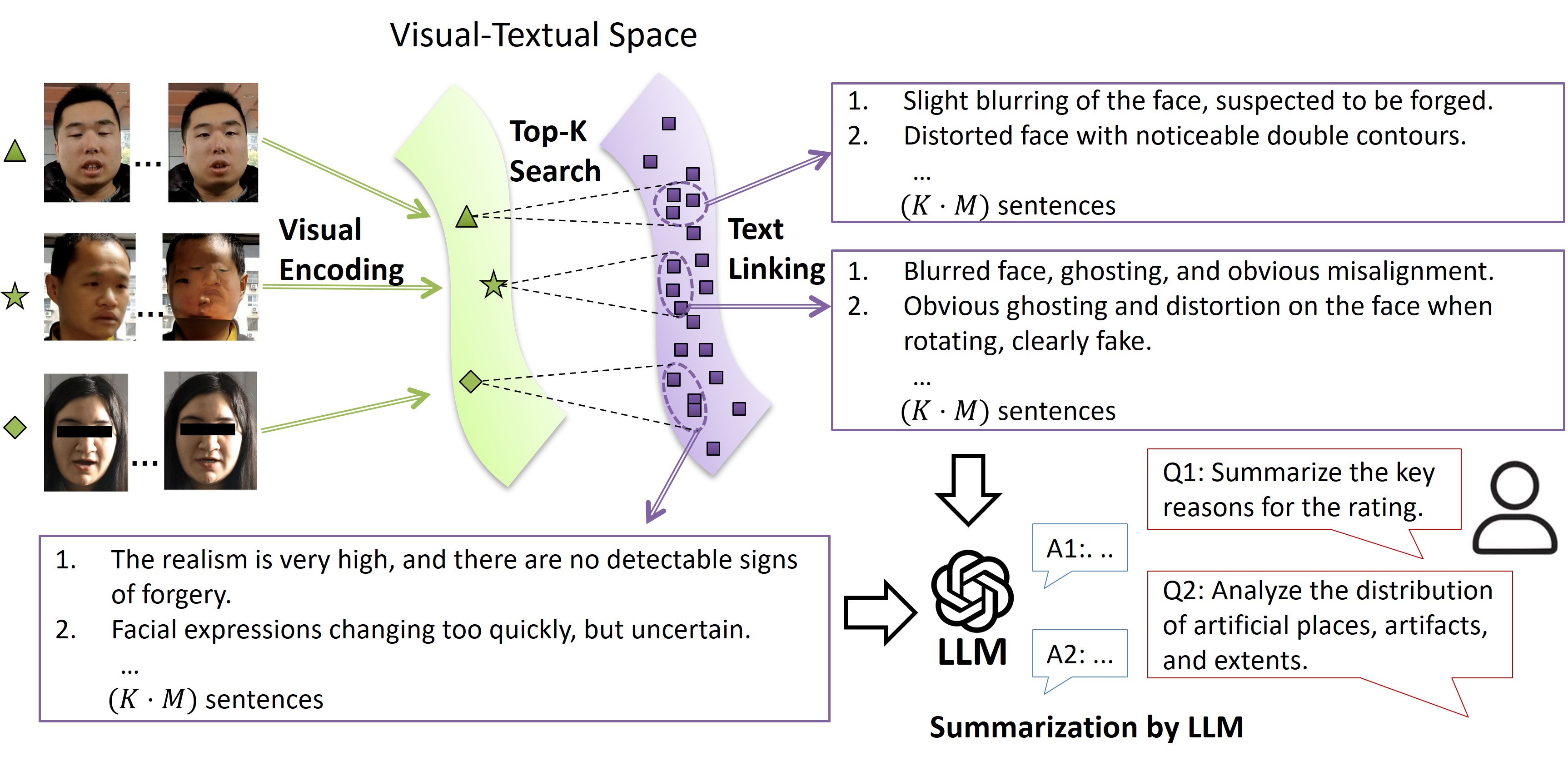}}
\caption{Illustration of the explanation procedure based on relevant sentence search in DA-CLIP's visual-textual space.}
\label{fig_explain}
\end{figure}
During inference, the visual branch can work alone to obtain photorealism assessment results. 
Apart from score prediction, we also show that explainability can be simply  achieved by leveraging the learned visual-textual alignment space, as shown in Fig. \ref{fig_explain}. Since we impose cross-modal similarity constraint during training, ideally, the visual feature of a query video should be most close to the features of relevant textual descriptions. Thus, for any query video, it first goes through the visual encoder to obtain the visual feature, and the top-$K$ closest textual representations of the training set videos are retrieved. This links the video to $K\cdot M$ most relevant description sentences, since each textual representation is the feature mean of all $M$ sentences describing the same video. These relevant sentences can serve as the raw explanation material for the predicted photorealism score of the query video. Further more, a Large Language Model (LLM) can be employed to summarize key information from the raw textual material in response to any human input questions. This is especially helpful since the number of sentences $K\cdot M$ is often overwhelmingly large for efficient overall comprehension ($M$ is 92 on average). We used the ChatGPT-o3 LLM in this work.

\section{Baseline Methods}
Apart from the proposed DA-CLIP method, we explore four types of baseline methods for benchmarking the deepfake photorealism assessment performance. They are hand-crafted feature based, deep feature based, finetuning based, and VLM based methods. The first two kinds only train a regression model on these features, as detailed in \cite{Sun2023VisualVideos}. The third kind finetunes some pretrained models for better adaption on the new task. The last kind adopts recent large vision language models for IQA and VQA. The methods we evaluate are summarized in Table \ref{tab_methods}. 
These baseline methods are briefly discussed in the Related Work \ref{sec-Related-IQA}, and we elaborate on some implementation details in the following. For complete understanding of these methods, we refer to their original papers.
\begin{table}[htb]
    \centering
    \caption{\centering Summary of all tested photorealism assessment methods. Some selected feature dimensions are in a range since multiple experiments are run.}
    \scriptsize
    \begin{tabular}{rcccc}
    \hline
        \textbf{Method} & \textbf{Type} & \multicolumn{2}{c}{\textbf{Feature dimension}} & \textbf{Pre-training data} \\
        \cmidrule(lr){3-4}
        ~ & ~  & \textbf{Original} & \textbf{Selected} & ~ \\ 
        \hline
        \textbf{BRISQUE\cite{Mittal2012No-referenceDomain}} & Hand-crafted & 72 & / & /  \\
        \textbf{GM-LOG\cite{Xue2014BlindFeatures}} & Hand-crafted & 80 & / & / \\
        \textbf{HIGRADE\cite{Kundu2017No-referencePictures}} & Hand-crafted & 432 & / & / \\
        \textbf{TLVQM\cite{Korhonen2019Two-levelAssessment}} & Hand-crafted &  75 & / & / \\
        \textbf{V-BLIINDS\cite{Saad2014BlindQuality}} & Hand-crafted &  46 & / & / \\
        \textbf{VIDEVAL\cite{Tu2021UGC-VQA:Content}} & Hand-crafted &  705 & 120$\sim$480 & / \\
        \hline
        \textbf{ResNet50\cite{He2016DeepRecognition}} & Deep feature &  4096 & 200$\sim$260 &  ImageNet \\
        \textbf{VGG-Face\cite{Parkhi2015DeepRecognition}} & Deep feature &  8192 & 320$\sim$480 & VGG-Face \\
        \textbf{DFGC-1st\cite{DFGC-2022Track}} & Deep feature &  4096 & 220$\sim$300 & Deepfake datasets \\
        \hline
        \textbf{OPDAI\cite{Peng2023DFGC-VRA:Assessment}} & Finetuning & 1536 & / & DFDC deepfake  \\
        \textbf{HUST\cite{Peng2023DFGC-VRA:Assessment}} & Finetuning & 768 & / & ImageNet  \\
        \textbf{UNILJ\cite{Peng2023DFGC-VRA:Assessment}} & Finetuning & 4096 & / & Deepfake datasets  \\
        \textbf{Patch-VQ\cite{Ying2021Patch-VQ:Problem}} & Finetuning & 4096 & / & LSVQ  \\
        \textbf{Fast-VQA\cite{Wu2022FAST-VQA:Sampling}} & Finetuning & 768 & / & Kinetics-400  \\
        \hline
        \textbf{Q-Align\cite{Wu2024Q-ALIGN:Levels}}     & VLM   & 4096   & /     & Multi-modal data  \\ 
        \textbf{DeQA-Score\cite{You2025TeachingDistribution}}   & VLM   & 4096   & /     & Multi-modal data  \\ 
        \textbf{CLIP-IQA \cite{Wang2023ExploringImages}}     & VLM   & 512   & /     & Multi-modal data  \\ 
        \textbf{DA-CLIP} & VLM & 512 & / & DFDC deepfake  \\
        \hline
    \end{tabular}
    \label{tab_methods}
\end{table}
\subsection{Hand-crafted and Deep Feature based Methods} \label{subsec_featurebased}
We first explore some handcrafted IQA features including BRISQUE \cite{Mittal2012No-referenceDomain}, GM-LOG \cite{Xue2014BlindFeatures}, and HIGRADE \cite{Kundu2017No-referencePictures}, and also some handcrafted VQA features, namely TLVQM \cite{Korhonen2019Two-levelAssessment}, V-BLIINDS\cite{Saad2014BlindQuality}, and VIDEVAL \cite{Tu2021UGC-VQA:Content}.
Besides the hand-crafted features, we also test features from pretrained deep models. These include the ResNet50 \cite{He2016DeepRecognition} model for object recognition, the VGG-Face \cite{Parkhi2015DeepRecognition} model for face recognition, and the DFGC-1st \cite{DFGC-2022Track, Peng2022DFGCCompetition} model for deepfake detection. 
We also use the feature selection method \cite{Tu2021UGC-VQA:Content} to reduce their feature dimension.


%
\subsection{Finetuning based Methods}
The top-3 methods from our competition work \cite{Peng2023DFGC-VRA:Assessment} are tested, i.e., the OPDAI, the HUST, and the UNILJ methods, which are all spacial networks, and we also test two spatial-temporal networks, i.e. Patch-VQ \cite{Ying2021Patch-VQ:Problem} and Fast-VQA \cite{Wu2022FAST-VQA:Sampling}. They are all finetuned on our deepfake photorealism assessment dataset. 

\textbf{The OPDAI method} employs the Swin-transformer v2 (\texttt{swinv2\_large\_window12to16\_192to256\_22kft1k}, 197M-parameters) \cite{Liu2022SwinResolution}. It is first pretrained on the DFDC deepfake detection dataset \cite{Dolhansky2020TheDataset}, and then finetuned on our photorealism assessment data. The finetuning minimizes two losses, i.e., the Norm-in-norm loss and the KL-divergence loss, as in Eqn. \ref{Eqn_OPDAI}.
For inference, three frames of a video are used for frame-level photorealism prediction and then averaged. 

\textbf{The HUST method} in this work is a simplified version of that in \cite{Peng2023DFGC-VRA:Assessment}, where we only train one model instead of five for ensemble. The base model is a ConvNeXt \cite{Liu2022A2020s} pretrained on the ImageNet dataset (\texttt{convnext\_tiny\_384\_in22ft1k}, 29M-parameters). 
The training loss is a combination of Mean Absolute Error (MAE) loss,  Pearson Linear Correlation Coefficient (PLCC) loss \cite{Wu2022FAST-VQA:Sampling} and a modified pair-wise ranking loss \cite{Wen2021AAssessment}. For inference, the video-level score is obtained by averaging 20 frame-level scores. 

\textbf{The UNILJ method} is also simplified from \cite{Peng2023DFGC-VRA:Assessment}, where we only train one model instead of two for ensemble. The base model is a ConvNeXt (\texttt{convnext\_xlarge\_384}, 350M-parameters) trained on a collection of 9 deepfake  datasets \cite{Peng2022DFGCCompetition}. 
The training loss is the Root of Mean Squared Error (RMSE) loss. 
For inference, the video-level prediction is averaged from predictions on 10 clips randomly selected from the video.

\textbf{Patch-VQ and Fast-VQA} are two spatial-temporal based models designed for VQA. Specifically, Patch-VQ \cite{Ying2021Patch-VQ:Problem} employs two-stream networks to extract both spatial feature (PaQ-2-PiQ) and temporal feature (3D ResNet-18), followed by local-to-global space-time feature pooling and finally temporal regression (54.2M parameters in total). We use the model pretrained on the LSVQ video quality assessment dataset. Since our finetuning dataset does not have local v-patch annotations, we use the Patch-VQ without v-patch version, and the finetuning loss is L1 loss. Then, Fast-VQA \cite{Wu2022FAST-VQA:Sampling} proposes the Fragment Attention Network (FANet) based on video swin-transformer (27.5M paprameters) and takes sampled fragments as input for better efficiency. The model is pretrained on the Kinetics-400 action recognition dataset, and the finetuning loss is PLCC loss. For these two methods, we pre-process the dataset using temporally aligned per-frame facial cropping to preserve temporal information.
\subsection{VLM based Methods}
We choose the Q-Align \cite{Wu2024Q-ALIGN:Levels}, the DeQA-Score \cite{You2025TeachingDistribution}, and the CLIP-IQA/CLIP-IQA+ \cite{Wang2023ExploringImages} which are widely compared VLM methods designed for assessing image and video quality. 

\textbf{Q-Align and  DeQA-Score.} 
The input prompt format is \textit{``Can you evaluate the photorealism of the video?"}, and the facial images are encoded and input to the large language model together with the textual embedding of the prompt. The expected output format is \textit{``The photorealism of the video is $\langle level \rangle$"}, where the $\langle level \rangle$ token is selected from \textit{\{``bad", ``poor", ``fair", ``good", ``excellent"\}}.
We tested two VLM backbones for these methods, i.e. the mPLUG-Owl2 model (8.2B-parameters) \cite{Ye2024Mplug-owl2:Collaboration} and the InternVL2.5-8B model (8B-parameters) \cite{Chen2024ExpandingScaling}.
At inference, a continuous photorealism score is obtained by weighting the $\langle level \rangle$ digits with their softmax probability.


\textbf{CLIP-IQA and CLIP-IQA+}. 
We use \textit{``High photorealism face images."} and \textit{``Low photorealism face images."} as the antonym prompts 
for this method. For an input frame, the visual feature 
is extracted by CLIP's visual branch, 
and the final photorealism score is obtained by softmax over its similarities with the antonym pair.
The backbones in CLIP-IQA/CLIP-IQA+ are ResNet-50 (25.6M-parameters) in the visual branch and a 12-layer 512-wide Transformer in the textual branch (63M-parameters).

\section{Experiments}
\subsection{Evaluation Settings}
\begin{figure}[ht]
\centering
\centerline{\includegraphics[width=0.35\textwidth]{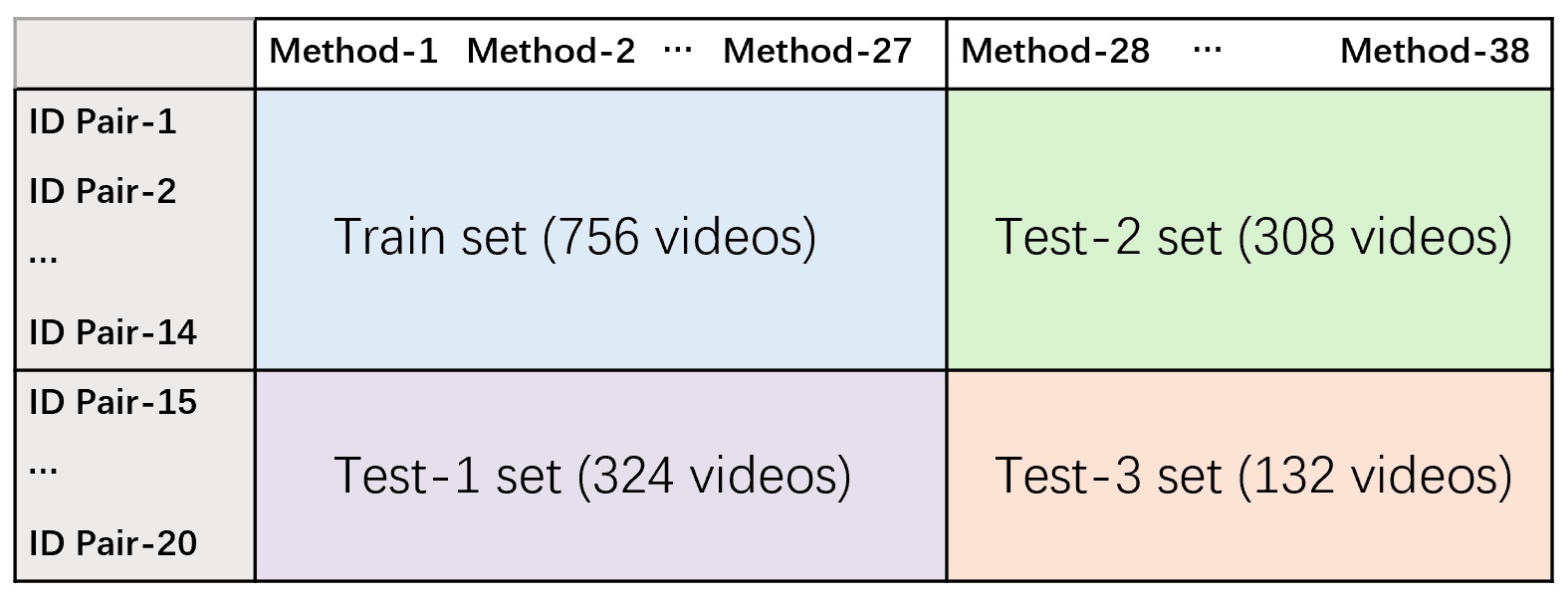}}
\caption{The splitting method of train and test sets for experimental evaluation.}
\label{fig_splits}
\end{figure}

The train-test splitting method is shown in Fig. \ref{fig_splits}. In our dataset, there are 20 pairs of captured actors, each pair is processed by 35 face-swap methods, and each person also has 3 real videos. We treat them  as 38 methods and split the dataset by methods and by actor IDs to obtain one train set and three test sets. This splitting method creates disjoint IDs and/or face-swap methods in train and test sets, which provides a challenging evaluation setting. To obtain more stable evaluation results, we repeat the train and test process 10 times using different splits to obtain the average performance. 

As for the evaluation metric, we use two metrics from the image and video quality assessment literature: Pearson Linear Correlation Coefficient (PLCC) and Spearman's Rank-order Correlation Coefficient (SRCC) to respectively evaluate the linearity and monotonicity of prediction with respect to the groundtruth. The two metrics are both in the range of $[-1, 1]$, higher the better. We calculate PLCC and SRCC on each individual test set and also average them to reflect the overall performance. 

The training and validation of the fine-tuning and VLM based methods are conducted on a 80\%-20\% split of the training set, the models are trained for 30 epochs, and finally the best checkpoint on the validation set is selected for testing. The optimizer uses AdamW, the initial learning rate is set to $1\times10^{-4}$. 
For training these methods, the video dataset is processed to extract one frame in every 6 frames, and the facial area is detected and cropped for the training and testing.
\subsection{Performance Comparison}
The results of compared methods are shown in Table \ref{tab_Results}. Here, \textit{PLCC1} represents the PLCC metric on the Test-1 set, so on and so forth, \textit{PLCC-avr} represents the average of PLCC metrics on all the three test sets, and \textit{avr} represents the final average of \textit{PLCC-avr} and \textit{SRCC-avr}. 
\begin{table*} [th] 
\centering
\caption{Comparison of different methods for the DREAM task. Results are in the \textit{mean$\pm$std} format obtained over 10 independent runs. \colorbox{mygray}{Gray background} represents quality assessment methods using hand-crafted features, \colorbox{mycyan}{cyan background} represents methods using deep features, \colorbox{mypink}{pink background} represents deep finetuning methods, and \colorbox{mylime}{lime background} represents VLM methods. The \textbf{best}, \underline{second}, and \textit{third} performances in each column are marked.}  \label{tab_Results}
\renewcommand{\arraystretch}{1.0}
\renewcommand{\arraystretch}{1.3}
\begin{threeparttable}
\scalebox{0.9}{%
\begin{tabular}{|c|c|c|c|c|c|c|c|c|c|}
\hline
Method     & PLCC1 $\uparrow$      & PLCC2 $\uparrow$      & PLCC3 $\uparrow$   & SRCC1 $\uparrow$  & SRCC2 $\uparrow$  & SRCC3 $\uparrow$  & PLCC-arv $\uparrow$   &SRCC-avr $\uparrow$   &avr $\uparrow$    \\
\hline
\rowcolor{mygray}
BRISQUE\cite{Mittal2012No-referenceDomain}  & \makecell{0.286$\pm$0.107}  & \makecell{0.478$\pm$0.144} 
& \makecell{0.225$\pm$0.085}   & \makecell{0.287$\pm$0.090}   &\makecell{0.580$\pm$0.103}
& \makecell{0.247$\pm$0.084}   & \makecell{0.330$\pm$0.112}   &\makecell{0.371$\pm$0.092}
& \makecell{0.350$\pm$0.102}  \\
\hline
\rowcolor{mygray}
GM-LOG\cite{Xue2014BlindFeatures}  & \makecell{0.455$\pm$0.072}  & \makecell{0.470$\pm$0.116} 
& \makecell{0.346$\pm$0.110}   & \makecell{0.487$\pm$0.077}   &\makecell{0.537$\pm$0.096}
& \makecell{0.401$\pm$0.105}   & \makecell{0.423$\pm$0.099}   &\makecell{0.475$\pm$0.092}
& \makecell{0.449$\pm$0.096}  \\
\hline
\rowcolor{mygray}
HIGRADE\cite{Kundu2017No-referencePictures}  & \makecell{0.426$\pm$0.060}  & \makecell{0.511$\pm$0.186} 
& \makecell{0.275$\pm$0.142}   & \makecell{0.437$\pm$0.065}   &\makecell{0.593$\pm$0.142}
& \makecell{0.310$\pm$0.151}   & \makecell{0.404$\pm$0.130}   &\makecell{0.446$\pm$0.119}
& \makecell{0.425$\pm$0.125}  \\
\hline
\rowcolor{mygray}
TLVQM\cite{Korhonen2019Two-levelAssessment}  & \makecell{0.525$\pm$0.073}  & \makecell{0.691$\pm$0.108} 
& \makecell{0.459$\pm$0.134}   & \makecell{0.466$\pm$0.071}   &\makecell{0.692$\pm$0.074}
& \makecell{0.403$\pm$0.118}   & \makecell{0.558$\pm$0.105}   &\makecell{0.520$\pm$0.088}
& \makecell{0.539$\pm$0.097}  \\
\hline
\rowcolor{mygray}
V-BLIINDS\cite{Saad2014BlindQuality}  & \makecell{0.531$\pm$0.059}  & \makecell{0.709$\pm$0.106} 
& \makecell{0.515$\pm$0.103}   & \makecell{0.433$\pm$0.077}   &\makecell{0.678$\pm$0.109}
& \makecell{0.440$\pm$0.099}   & \makecell{0.585$\pm$0.089}   &\makecell{0.517$\pm$0.095}
& \makecell{0.551$\pm$0.092}  \\
\hline
\rowcolor{mygray}
VIDEVAL\cite{Tu2021UGC-VQA:Content}  & \makecell{0.621$\pm$0.053}  & \textit{\makecell{0.799$\pm$0.067}} 
& \makecell{0.595$\pm$0.092}   & \makecell{0.560$\pm$0.056}   &\makecell{0.761$\pm$0.104}
& \makecell{0.527$\pm$0.103}   & \makecell{0.672$\pm$0.071}   &\makecell{0.616$\pm$0.087}
& \makecell{0.644$\pm$0.079}  \\
\hline
\rowcolor{mycyan}
ResNet50\cite{He2016DeepRecognition}  & \makecell{0.371$\pm$0.073}  & \makecell{0.672$\pm$0.113} 
& \makecell{0.299$\pm$0.139}   & \makecell{0.357$\pm$0.063}   &\makecell{0.663$\pm$0.092}
& \makecell{0.314$\pm$0.131}   & \makecell{0.447$\pm$0.109}   &\makecell{0.445$\pm$0.095}
& \makecell{0.446$\pm$0.102}  \\
\hline
\rowcolor{mycyan}
VGG-face\cite{Parkhi2015DeepRecognition}  & \makecell{0.237$\pm$0.092}  & \makecell{0.631$\pm$0.092} 
& \makecell{0.220$\pm$0.085}   & \makecell{0.200$\pm$0.067}   &\makecell{0.625$\pm$0.095}
& \makecell{0.208$\pm$0.073}   & \makecell{0.363$\pm$0.089}   &\makecell{0.344$\pm$0.078}
& \makecell{0.353$\pm$0.084}  \\
\hline
\rowcolor{mycyan}
DFGC1st\cite{DFGC-2022Track}  & \makecell{0.727$\pm$0.070}  & \makecell{0.755$\pm$0.074} 
& \makecell{0.616$\pm$0.078}   & \makecell{0.680$\pm$0.073}   &\makecell{0.740$\pm$0.085}
& \makecell{0.576$\pm$0.083}   & \makecell{0.699$\pm$0.074}   &\makecell{0.665$\pm$0.080}
& \makecell{0.682$\pm$0.077}  \\
\hline
\rowcolor{mypink}
HUST\cite{Peng2023DFGC-VRA:Assessment}  & \makecell{0.629$\pm$0.088 }  & \makecell{0.608$\pm$0.088 } 
& \makecell{0.527$\pm$0.099}   & \makecell{0.634$\pm$0.081 }   &\makecell{0.634$\pm$0.061}
& \makecell{0.545$\pm$0.085}   & \makecell{0.588$\pm$0.058 }   &\makecell{0.604$\pm$0.032}
& \makecell{0.596$\pm$0.043}  \\
\hline
\rowcolor{mypink}
UNILJ\cite{Peng2023DFGC-VRA:Assessment}  & \textit{\makecell{0.797$\pm$0.063}}  & \makecell{0.749$\pm$0.077} 
&\textit{\makecell{0.620$\pm$0.089}}   & \textit{\makecell{0.747$\pm$0.060}}   &\makecell{0.718$\pm$0.098}
& \textit{\makecell{0.582$\pm$0.113}}   & \textit{\makecell{0.722$\pm$0.044}}   &\textit{\makecell{0.682$\pm$0.059}}
& \textit{\makecell{0.702$\pm$0.050}}  \\
\hline
\rowcolor{mypink}
OPDAI\cite{Peng2023DFGC-VRA:Assessment}  & \underline{\smash{\makecell{0.832$\pm$0.049}}}  & \underline{\smash{\makecell{0.835$\pm$0.084}}}
& \underline{\smash{\makecell{0.738$\pm$0.116}}}   & \underline{\smash{\makecell{0.772$\pm$0.063}}}   &\textbf{\makecell{0.818$\pm$0.073}}
& \underline{\smash{\makecell{0.697$\pm$0.107}}}   & \underline{\smash{\makecell{0.802$\pm$0.065}}}   &\underline{\smash{\makecell{0.762$\pm$0.053}}}
& \underline{\smash{\makecell{0.782$\pm$0.056}}}  \\
\hline
\rowcolor{mypink}
Patch-VQ\cite{Ying2021Patch-VQ:Problem}  & \makecell{0.483$\pm$0.084}  & \makecell{0.617$\pm$0.115} 
& \makecell{0.317$\pm$0.129}   & \makecell{0.444$\pm$0.080}   &\makecell{0.599$\pm$0.106}
& \makecell{0.303$\pm$0.157}   & \makecell{0.472$\pm$0.073}   &\makecell{0.449$\pm$0.084}
& \makecell{0.460$\pm$0.077}  \\
\hline
\rowcolor{mypink}
Fast-VQA\cite{Wu2022FAST-VQA:Sampling}  & \makecell{0.782$\pm$0.050}  & \makecell{0.739$\pm$0.086} 
& \makecell{0.596$\pm$0.132}   & \makecell{0.708$\pm$0.047}   &\makecell{0.688$\pm$0.088}
& \makecell{0.517$\pm$0.139}   & \makecell{0.706$\pm$0.071}   &\makecell{0.638$\pm$0.070}
& \makecell{0.672$\pm$0.068}  \\
\hline
\rowcolor{mylime}
Q-Align-m\cite{Wu2024Q-ALIGN:Levels}  & \makecell{0.227$\pm$0.075}  & \makecell{0.166$\pm$0.108} 
& \makecell{0.147$\pm$0.109}   & \makecell{0.265$\pm$0.081}   &\makecell{0.202$\pm$0.106}
& \makecell{0.212$\pm$0.099}   & \makecell{0.180$\pm$0.051}   &\makecell{0.226$\pm$0.051}
& \makecell{0.203$\pm$0.050}  \\
\hline
\rowcolor{mylime}
Q-Align-i\cite{Wu2024Q-ALIGN:Levels}  & \makecell{0.685$\pm$0.088}  & \makecell{0.697$\pm$0.090} 
& \makecell{0.561$\pm$0.103}   & \makecell{0.661$\pm$0.068}   &\makecell{0.698$\pm$0.067}
& \makecell{0.563$\pm$0.079}   & \makecell{0.648$\pm$0.068}   &\makecell{0.641$\pm$0.052}
& \makecell{0.644$\pm$0.057}  \\
\hline
\rowcolor{mylime}
DeQA-m\cite{You2025TeachingDistribution}  & \makecell{0.238$\pm$0.071}  & \makecell{0.177$\pm$0.103} 
& \makecell{0.164$\pm$0.114}   & \makecell{0.275$\pm$0.081}   &\makecell{0.210$\pm$0.105}
& \makecell{0.217$\pm$0.101}   & \makecell{0.193$\pm$0.053}   &\makecell{0.234$\pm$0.050}
& \makecell{0.213$\pm$0.051}  \\
\hline
\rowcolor{mylime}
DeQA-i\cite{You2025TeachingDistribution}  & \makecell{0.756$\pm$0.052} 
& \makecell{0.777$\pm$0.090}   & \makecell{0.598$\pm$0.111}   &\makecell{0.697$\pm$0.060}
&\textit{\makecell{0.765$\pm$0.088}}   & \makecell{0.558$\pm$0.092}   &\makecell{0.710$\pm$0.058}
& \makecell{0.674$\pm$0.058} & \makecell{0.692$\pm$0.056} \\
\hline
\rowcolor{mylime}
CLIP-IQA\cite{Wang2023ExploringImages}  & \makecell{0.026$\pm$0.075}  & \makecell{-0.006$\pm$0.043} 
& \makecell{0.022$\pm$0.113}   & \makecell{0.022$\pm$0.070}   &\makecell{0.004$\pm$0.052}
& \makecell{0.048$\pm$0.101}   & \makecell{0.014$\pm$0.058}   &\makecell{0.025$\pm$0.050}
& \makecell{0.019$\pm$0.053}  \\
\hline
\rowcolor{mylime}
CLIP-IQA+\cite{Wang2023ExploringImages}  & \makecell{0.082$\pm$0.091}  & \makecell{0.125$\pm$0.117} 
& \makecell{0.086$\pm$0.086}   & \makecell{0.092$\pm$0.091}   &\makecell{0.116$\pm$0.136}
& \makecell{0.090$\pm$0.096}   & \makecell{0.097$\pm$0.050}   &\makecell{0.099$\pm$0.059}
& \makecell{0.098$\pm$0.054}  \\
\hline
\rowcolor{mylime}
DA-CLIP  & \textbf{\makecell{0.842$\pm$0.034}}  & \textbf{\makecell{0.872
$\pm$0.049}} 
& \textbf{\makecell{0.856$\pm$0.061}}   & \textbf{\makecell{0.784
$\pm$0.027}}   &\underline{\smash{\makecell{0.817
$\pm$0.037}}}
& \textbf{\makecell{0.794$\pm$0.045}}   & \textbf{\makecell{0.857
$\pm$0.028}}   &\textbf{\makecell{0.798
$\pm$0.021}}
& \textbf{\makecell{0.827$\pm$0.021}}  \\
\hline
\rowcolor{mylime}
DA-CLIP-T \tnote{1}  & \makecell{0.977$\pm$0.004}  & \makecell{0.976$\pm$0.010} 
& \makecell{0.975$\pm$0.011}   & \makecell{0.971
$\pm$0.005}   &\makecell{0.969$\pm$0.010}
& \makecell{0.961$\pm$0.019}   & \makecell{0.976$\pm$0.006}   &\makecell{0.967$\pm$0.008}
& \makecell{0.971$\pm$0.007}  \\
\hline
\end{tabular} 
} 
\begin{tablenotes}
\item[1] DA-CLIP-T denotes its textual branch that predicts MOS based on textual descriptions, hence its performance is extraordinary. It is just listed as a \\reference for the other visual based methods.
\end{tablenotes}
\end{threeparttable}
\end{table*}

In the group of hand-crafted feature based methods, VQA methods (i.e. TLVQM, V-BLIINDS, and VIDEVAL) surpass IQA methods (i.e. BRISQUE, GM-LOG, and HIGRADE), implying the effectiveness of motion features in deepfake video photorealism assessment. VIDEVAL achieves the best performance in this group, benefiting from its ensembled features and the feature selection process that makes it more adapted on the DREAM task. 

For the group of deep feature based methods, the performance is affected by the pre-training tasks. The ResNet50 feature is for general object recognition, the VGG-face feature is for facial identity recogntion, and they both obtain results that are no better than hand-crafted feature based methods. On the contrary, the DFGC1st feature, which is originally trained for deepfake detection, achieves far better result, and it is even better than VIDEVAL. This may be attributed to the closer internal relation between deepfake detection and deepfake photorealism assessment. Although the two tasks are distinct, they may have overlap in focusing on subtle micro characteristics of facial videos. 

In the group of fine-tuning based methods, both UNILJ and OPDAI surpass the deep feature based method DFGC1st, and they respectively achieves the third and second best overall performance in all evaluated methods. Notably, the OPDAI method achieves average result of 0.782±0.056, which is
10 points higher than the DFGC1st performance. Their results verify the importance of proper fine-tuning for the DREAM task. More ablation study on these fine-tuning based methods are conducted in Subsection \ref{subsec_ablation} to show the main components leading to their effectiveness.
As for the spatial-temporal methods Patch-VQ and Fast-VQA, although they explicitly model the temporal information, their performance did not surpass the DFGC1st method, implying the importance of using more relevant pretraining dataset.

We further compare the VLM based methods, which are more recent approaches investigated in the IQA/VQA field. Q-Align and DeQA-Score are both based on finetuning multi-modal large language models to output language tokens that indicate the photorealism level. Their performance is very dependent on the adopted back-bone large models, where we tested the mPLUG-Owl2 (-m) and the
InternVL2.5-8B (-i). Results show the InternVL2.5-8B (-i) versions consistently surpass the counterparts, which may be attributed to its stronger or more related pretrained capacity. Notably, the DeQA-Score-i method has the fourth best overall performance, making it a close match to the UNILJ method. The other kind of VLM based method, i.e., CLIP-IQA and CLIP-IQA+, on the contrary have the lowest performance. This is not surprising though, since these methods fix the entire model weights and CLIP-IQA+ only tunes the input prompt, making them theoretically more close to feature based methods. 

Finally, we see that the proposed DA-CLIP method achieves the best overall performance. Since its architecture adopts the same visual backbone  and MOS regression loss as in the OPDAI method, we attribute its main improvement to the incorporation of description alignment in the multi-modal training process. The description alignment helps the visual branch to learn more fine-grained and effective representations that are useful in DREAM, and we conduct more in-depth analyses of it in Subsection \ref{subsec-explanability} and \ref{subsec-explaination}. In the table, we also listed the performance of the textual branch of DA-CLIP, i.e. DA-CLIP-T, for reference, though it is not directly comparable with the other visual based methods. Its prediction has near perfect agreement with the groundtruth MOS, implying that the textual descriptions contain more indicative and relevant information.

We also analyze the performance variations among the three test sets to see the impacting factors for generalization. It can be seen that the Test-3 set is the most difficult one, which has the lowest PLCC and SRCC for nearly all methods. This is because it has both disjoint IDs and disjoint deepfake creation methods that are different from the training set, making the generalization most difficult. By comparing the performances on Test-1 and Test-2 sets, it can be seen that Test-1 is more difficult for most methods in terms of PLCC and SRCC. It implies that different IDs make more challenges than different deepfake methods, as is the case for the current dataset.

\subsection{Effectiveness of Different Losses and Pretrainings} \label{subsec_ablation}
The analysis is conducted on the fine-tuning based methods OPDAI and UNILJ, which achieve the second and third best performance. Two aspects are analyzed, i.e., the effects of different loss functions and the effects of different kinds of pre-training data, since they are the most notable variations across different methods.

We first analyze the impact of loss functions using the OPDAI method. Individual and combined losses from fine-tuning based methods are tested. The results are shown in Table \ref{tab_Loss}. For single losses, RMSE beats MAE by a clear margin, and both NinN and KL loss clearly surpass RMSE, with the KL loss achieving the best performance in this case. The superiority of NinN and KL manifests the effectiveness of the normalization of scores in each batch before loss calculation. For combination of losses, since the Rank loss and PLCC loss cannot be used alone, we combine them with the RMSE loss. It shows that the PLCC loss is effective in further improving the performance of RMSE, while the Rank loss is not effective. It is also surprising to find that NinN combined with KL can impair some performance, although each loss alone is very effective. This result goes against the combined loss used in the original OPDAI method, but still the new best performance of OPDAI with KL loss is lower than the proposed DA-CLIP method.
\begin{table} [th] 
\centering
\caption{Analysis on the effectiveness of different losses on the OPDAI method. \textbf{Bold} and \underline{underlined} numbers respectively represent the best and second best results.}  \label{tab_Loss}
\renewcommand{\arraystretch}{1.0}
\renewcommand{\arraystretch}{1.3}
\scalebox{0.9}{%
\begin{tabular}{|c|c|c|c|}
\hline
Loss   & PLCC-arv $\uparrow$   & SRCC-avr $\uparrow$   & avr $\uparrow$    \\
\hline
MAE  & 0.624$\pm$0.045   &0.638$\pm$0.034 &0.631$\pm$0.037  \\
 \hline
RMSE  & 0.753$\pm$0.030   &0.697$\pm$0.039  &0.725$\pm$0.033 \\
 \hline
 NinN        &\underline{0.813$\pm$0.051}    &\underline{0.775$\pm$0.042}    &\underline{0.794$\pm$0.043}    \\
\hline
KL          &\textbf{0.829$\pm$0.069}    &\textbf{0.791$\pm$0.055}    &\textbf{0.810$\pm$0.061}    \\
\hline
RMSE+Rank   & 0.753$\pm$0.068   &0.699$\pm$0.049    &0.726$\pm$0.050    \\
 \hline
RMSE+PLCC   &0.809$\pm$0.051    &0.766$\pm$0.037    &0.788$\pm$0.040    \\
\hline
NinN+KL     &0.802$\pm$0.065    &0.762$\pm$0.053    &0.782$\pm$0.056    \\
\hline
\end{tabular} 
} 
\end{table}

The analysis of using different types of pretraining data is conducted on both the OPDAI and UNILJ methods while keeping their original losses, because they both pre-trained their models on deepfake detection datasets. The results are shown in Table \ref{tab_pretrain}. As can be seen, pretraining on Deepfake datasets clearly improves performance compared with pretraining on the more general ImageNet dataset. The OPDAI's average performance is improved by 3 points and the UNILJ's is improved by 9 points. This result is in-line with the comparison of feature based methods in Table \ref{tab_Results}, where the DFGC1st feature surpassed all the other features. It again emphasizes the underlying close relation between the DREAM task and the deepfake detection task. It should be noted that the deepfake detection datasets for pretraining do not have overlap with our photorealism assessment dataset.
\begin{table} [th] 
\centering
\caption{Analysis on the effectiveness of different pretraining data.}  \label{tab_pretrain}
\renewcommand{\arraystretch}{1.0}
\renewcommand{\arraystretch}{1.3}
\scalebox{0.9}{%
\begin{tabular}{|c|c|c|c|}
\hline
Method \& Pretraining   & PLCC-arv $\uparrow$   & SRCC-avr $\uparrow$   & avr $\uparrow$    \\
\hline
OPDAI on ImageNet  &0.774$\pm$0.076     &0.727$\pm$0.058    &0.750$\pm$0.064    \\
\hline
OPDAI on Deepfake  &0.802$\pm$0.065     &0.762$\pm$0.05     &0.782$\pm$0.056    \\
\hline
UNILJ on ImageNet  &0.631$\pm$0.066     &0.591$\pm$0.073    &0.611$\pm$0.067    \\
\hline
UNILJ on Deepfake  &0.722$\pm$0.044     &0.682$\pm$0.059    &0.702$\pm$0.050    \\
\hline
\end{tabular} 
} 
\end{table}
\subsection{Interpretability of DA-CLIP} \label{subsec-explanability}
In this subsection, we investigate the interpretability of the proposed DA-CLIP method given its very good performance. The model has a textual branch and a visual branch, and each can independently perform photorealism score regression. We first analyze each branch to see the contribution of textual or visual tokens in regressing MOS scores, respectively, and then cross-modal similarity in the feature space is examined.

\begin{figure}[th]
\centering
\centerline{\includegraphics[width=0.4\textwidth]{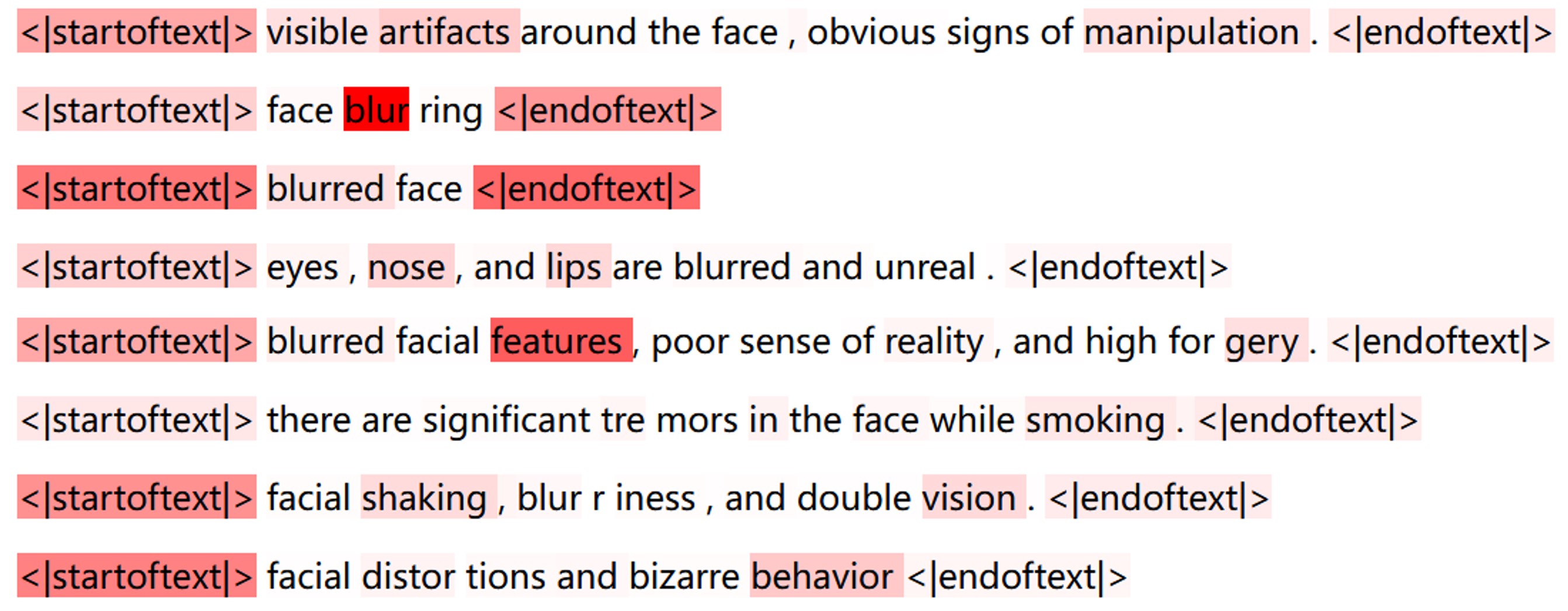}}
\caption{Illustration of part of the textual descriptions and their importance to the textual branch prediction, highlighted with different shades of red.}
\label{fig_textualCAM}
\end{figure}

\begin{figure}[th]
\centering
\centerline{\includegraphics[width=0.5\textwidth]{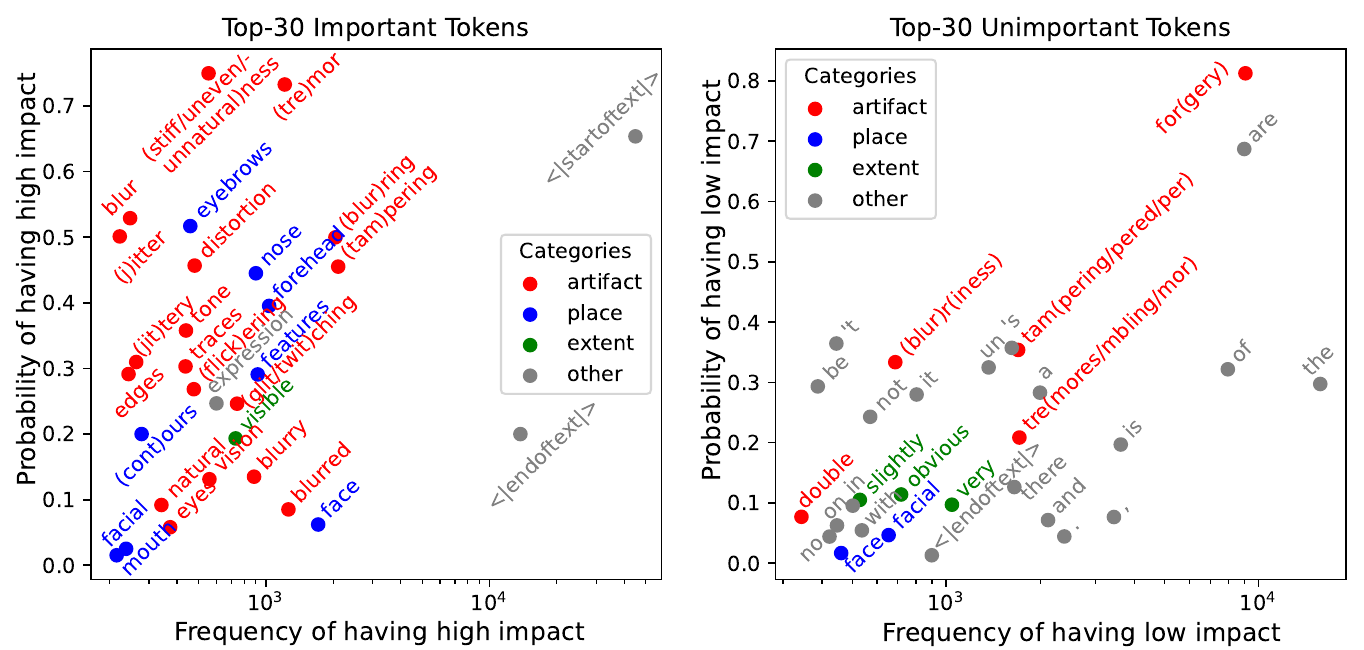}}
\caption{Top-30 most important (left) and most unimportant (right) textual tokens. Since many tokens are part of words, we complete them in parentheses for reading convenience.}
\label{fig_importantTokens}
\end{figure}

First, the contribution of each textual token in the textual branch model is analyzed. The contribution weights are calculated using the attentive class activation AttCAT method \cite{Qiang2022Attcat:Tokens}, which leverages encoded features, their gradients, and their attention weights to attribute output scores to input tokens. An example of the textual descriptions of a video and their importance is shown in Fig. \ref{fig_textualCAM}. We then show the top-30 most important and unimportant tokens to the textual prediction model in Fig. \ref{fig_importantTokens}, selected over all textual descriptions in the test sets. Here, the important tokens are selected as the top 10\% important ones for a video, and the unimportant tokens are the bottom 10\% ones. The frequency of each token appearing in the (un)important list is then counted, and the top-30 most frequent ones are shown. Along with the frequency, we also calculate the probability of these tokens to appear in the (un)important list by dividing their total counts. As shown in Fig. \ref{fig_importantTokens}, among the important tokens, there are 18 describing artifact, 8 describing places or locations, 1 describing extent, and 3 others. While for the unimportant tokens, it has 20 others, 5 artifacts, 3 extents, and 2 places. These statistics are reasonable, since artifacts and places are most directly related to the perception of photorealism, while other tokens like linking verbs and prepositions are not important. On the other hand, the extent tokens turn out to be not  important, which is surprising at first glance. We attribute this phenomenon to the inconsistent or even contradictory descriptions of extent among different annotators, which is actually quite normal since people often have different sense of an artifact's obviousness, but they tend to have more agreements on the existence of the artifact. The \textit{startoftext} and \textit{endoftext} tokens are important since the sentence-level features are taken from the \textit{endoftext} token and together they mark the length of a sentence. Finally, we note a few tokens or their complete words can appear in both the important and unimportant figures, e.g. \textit{tam(pering)} and \textit{tre(mor)}. This may be due to the splitting of one word to different parts in the tokenization process and can also be seen as a result of information redundancy when the model is given a large number of textual descriptions.

\begin{figure}[th]
\centering
\centerline{\includegraphics[width=0.5\textwidth]{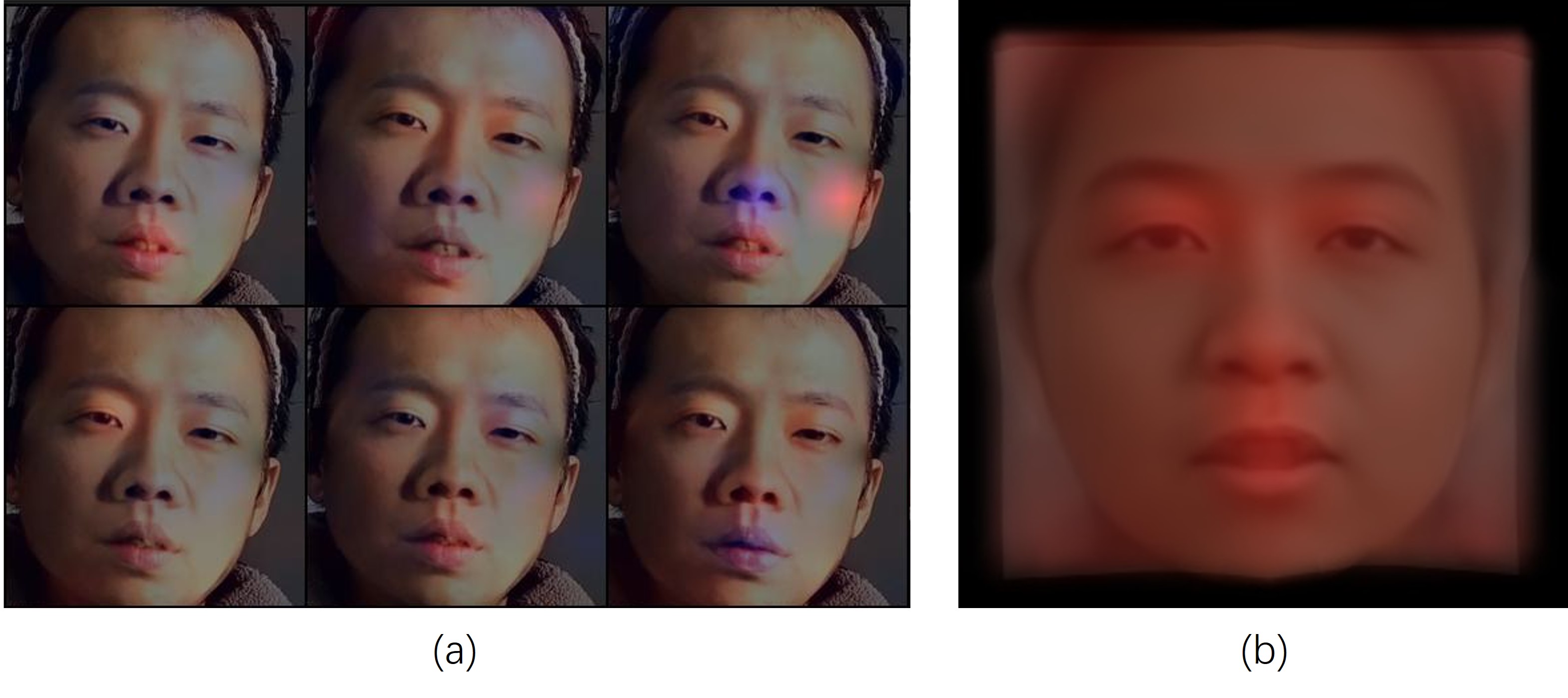}}
\caption{Sub-figure (a) shows important visual regions across a few sample frames of a video, and (b) is the overall importance map overlaid on the aligned average face of the test dataset.}
\label{fig_visualCAM}
\end{figure}

Then, we analyze the visual branch to see the important image locations that are important for the VRA prediction. This analysis also employs the AttCAT method \cite{Qiang2022Attcat:Tokens}, given that the visual branch is also based on the Transformer architecture. We treat each local patch as a visual token and each frame as a sentence, thus the video photorealism prediction can be attributed to each input location using AttCAT. The visualization result is shown in Fig. \ref{fig_visualCAM}, where in (a) we use red and blue colors to represent positive and negative impacts respectively, and in (b) the absolute values of these impacts are summed and averaged over all test videos. As can be seen from (a), our photorealism assessment model can focus on diverse different locations across different video frames, including mouth, teeth, eyes, nose, and borders. Since the VRA is a regression task, both positive- and negative- impacting areas are important for the prediction. Sub-figure (b) is the canonically morphed and aligned average face of all people in the test sets, overlaid with the average importance map. It shows that the visual branch model commonly resort to key facial features for photorealism assessment, and the background is also important, probably for being contrasted with by the facial features to better reveal blurriness and other artifacts.  

\begin{figure}[th]
\centering
\centerline{\includegraphics[width=0.3\textwidth]{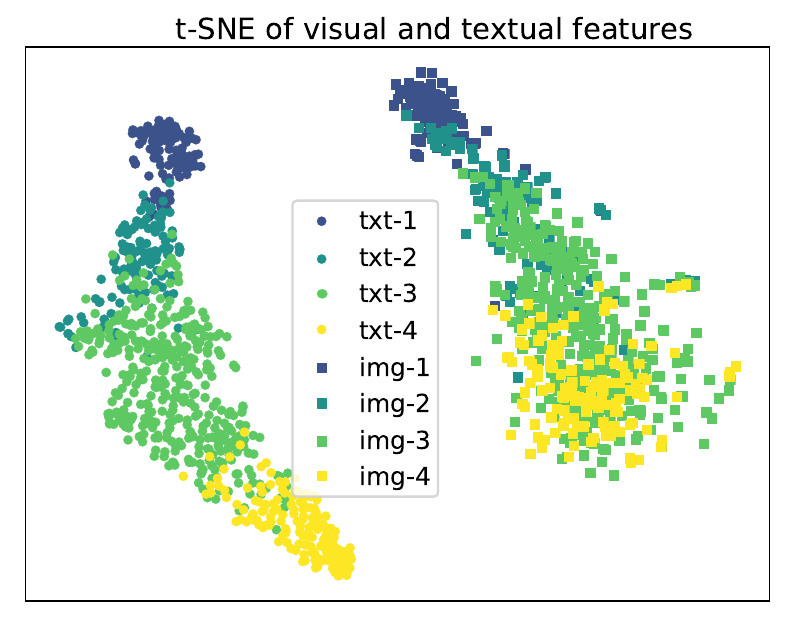}}
\caption{t-SNE visualization of visual (img) and textual (txt) features extracted on the test sets. The group 1, 2, 3, 4 represents videos with MOS scores in the range of [1, 2), [2, 3), [3, 4), and [4, 5), respectively.}
\label{fig_tSNE}
\end{figure}

\begin{figure}[th]
\centering
\centerline{\includegraphics[width=0.35\textwidth]{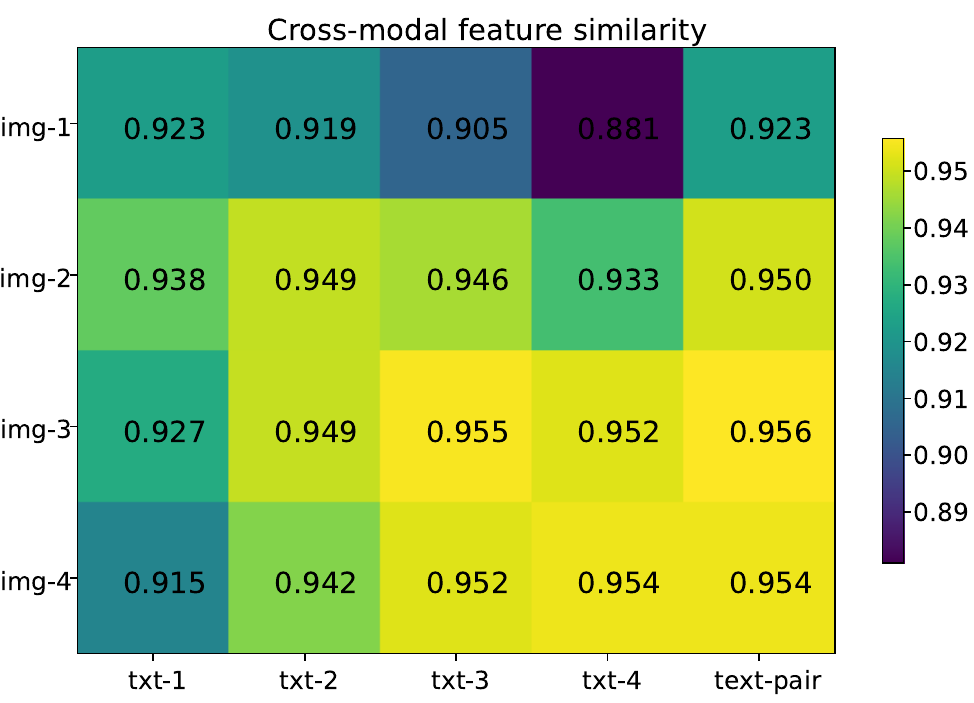}}
\caption{Average of cross-modal cosine similarity, calculated from independently sampled visual/textual features from each MOS group (the first four columns), or from correspondingly sampled pairs in each group (the last column).}
\label{fig_cross_sim}
\end{figure}

Finally, we show the cross-modal similarity of corresponding visual and textual features. The t-SNE plot visualizing their distribution in the shared representation space of CLIP is shown in Fig. \ref{fig_tSNE}. Although an explicit cross-modal similarity loss is imposed on the model, and we do observe a normal decrease of this loss during training, there is still a modality gap between visual and textual features.  This gap is caused by a combination of model initialization and contrastive learning optimization \cite{Liang2022MindLearning} and commonly observed in CLIP-based models. More importantly, we can see a smooth transition between features from adjacent score groups, more prominently in the textual modality. And corresponding groups from the two modalities are generally in parallel, and they are relatively closer to each other compared to those from a non-corresponding group. We then quantitatively verify this in Fig. \ref{fig_cross_sim}. The figure is calculated by independently sampling 10,000 pairs of textual and visual features from a combination of score groups and obtain their averaged cosine similarity. From the first four columns, we can see that the diagonal has the largest cross-modal similarity from the visual perspective. That is for the visual features from every score group, the closest textual features are from the same score group on average. Further comparing the fifth column with the diagonal, the exact corresponding textual feature is closer or at least equally close to a query visual feature compared to some general textual features from the same group. This verifies that the adapted CLIP model successfully learned the cross-modal similarity relationship that pulls corresponding pairs closer. However, due to the natural vagueness in human's textual description of photorealism, a clear-cut exceeding of corresponding pairs over other similar ones is not observed.  


\subsection{Textual-based Explanation by DA-CLIP} \label{subsec-explaination}
In the following, we validate the proposed explanation method that is based on relevant sentence search and LLM summarization.
Firstly, we ask LLM to summarize for each test video the 37-dimensional key categorical feature as in Fig. \ref{fig_category-hist}. It is regarded as a quantified form of the textual explanations, since it describes the place, artifact, extent, and their ratios, which can serve as a fine-grained assessment for the video. We conduct an evaluation on the test sets and use the categorical features summarized from their original annotations as the groundtruth. The average of Root Mean Square Error (RMSE) between the predicted category features and the groundtruth ones is evaluated and compared across different top-K searching strategies, and the result is shown in Table \ref{tab_top-K}. The random-1 strategy is shown as a reference, i.e., randomly selecting one textual feature from the training set and summarize the category features from its associated descriptions. As can be seen from the table, the top-K strategy clearly reduced explanation errors from the aspects of both overall and the individual group of key categories. With the increase of $K$, the explanation error further decrease, and top-11 is a good balance between accuracy and efficiency.

\begin{table} [th] 
\centering
\caption{The explanation error by RMSE ($\downarrow$) between predicted description categories and groundtruth.}  \label{tab_top-K}
\renewcommand{\arraystretch}{1.0}
\scalebox{0.9}{%
\begin{tabular}{|c|c|c|c|c|}
\hline
Strategy  & All$\uparrow$   & Place   & Artifact   & Extent    \\
\hline
Random-1  &0.099$\pm$0.001     &0.098$\pm$0.001    &0.084$\pm$0.001   &0.137$\pm$0.002    \\
\hline
Top-1  &0.083   &0.073  &0.079  &0.112  \\
\hline
Top-3  &0.070   &0.063  &0.065  &0.093  \\
\hline
Top-5  &0.066   &0.060  &0.061  &0.089  \\
\hline
Top-7  &0.065   &0.059  &0.059  &0.087  \\
\hline
Top-9  &0.064   &0.059  &0.058  &0.085  \\
\hline
Top-11  &\textbf{0.063}  &0.059  &0.058  &\textbf{0.085}  \\
\hline
Top-13  &\textbf{0.063}  &\textbf{0.058}  &\textbf{0.057}  &\textbf{0.085}  \\
\hline
\end{tabular} 
} 
\end{table}
\begin{figure*}[ht]
\centering
\centerline{\includegraphics[width=1.0\textwidth]{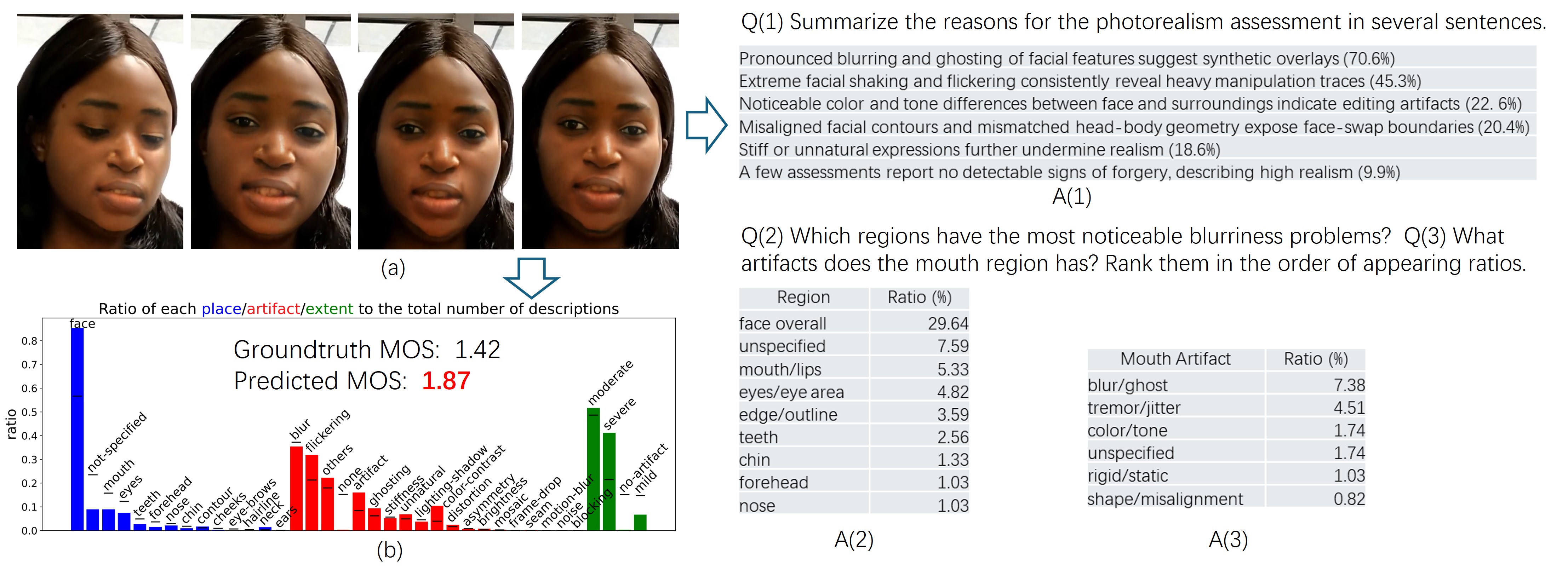}}
\caption{Demonstration of the DA-CLIP capability in deepfake photorealism assessment and explanation. (a) is the frames of an tested deepfake video, (b) is the predicted distribution of key description categories, and we also show three Q\&A scenarios by prompting ChatGPT-o3 providing the retrieved descriptions by our top-11 strategy.}
\label{fig_demo}
\end{figure*}
\begin{figure}[ht]
\centering
\centerline{\includegraphics[width=0.5\textwidth]{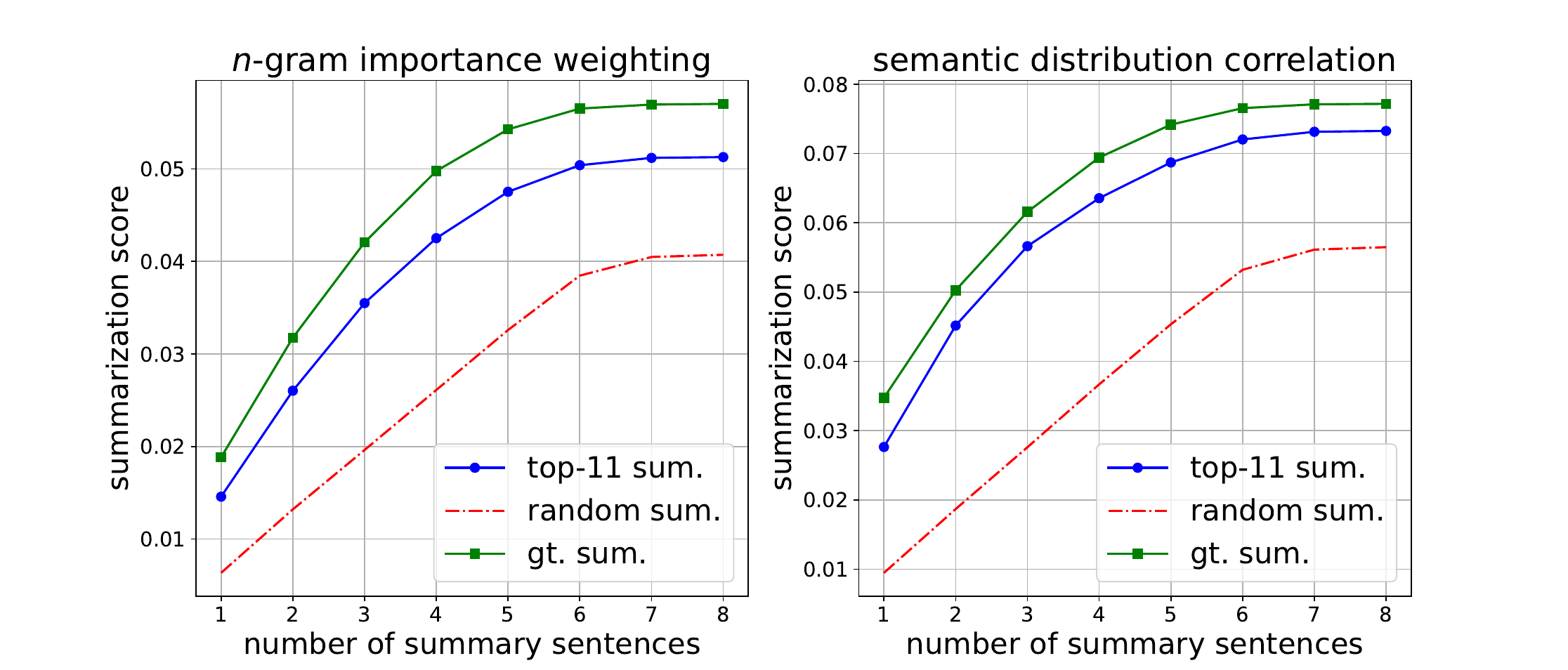}}
\caption{Summarization scores measured by two different metrics (both higher the better) with different number of summary sentences.}
\label{fig_sumScore}
\end{figure}

Lastly, we show a demonstration of the DA-CLIP capability in deepfake photorealism assessment and explanation in Fig. \ref{fig_demo}. The input deepfake video is a low-photorealism one with $1.42$ groundtruth MOS. It has very obvious flickering when viewed in video format (note the brightness change between the 2nd and the 3rd frames at the lower-right cheek for example), the mouth and teeth area is especially blurry and has stiff movements, and the splicing seam is noticeable at the lower contour. From (b), we can see that the predicted flickering ratio largely exceeds its average height (i.e., the one in Fig. \ref{fig_category-hist}). Other notable observations in (b) include people tend to give more descriptions on the whole face when the photorealism is quite low, and artifact and color-contrast problems are more prominent. As shown in Q\&A (1-3), we can further employ ChatGPT-o3 for more flexible and in-depth assessments, where we first provide the 975 descriptions retrieved by the top-11 strategy to ChatGPT, and then ask specific questions using carefully designed prompts. As observed from the ChatGPT answers, it reasonably summarized the key artifact types, analyzed the distribution of blurriness over facial regions, and analyzed the existence of different artifacts at the mouth area. These flexible interactions enable more in-depth and fine-grained insights for deepfake photorealism assessment. 

We further quantitatively evaluate the faithfulness of ChatGPT generated summarizations, i.e. Fig. \ref{fig_demo} A(1), in Fig. \ref{fig_sumScore}. Since no groundtruth human-made summarizations are available, we use two no-reference summarization metrics, which are the $n$-gram importance weighting based metric \cite{Gigant2024MitigatingMetrics} and the semantic distribution correlation (SDC) metric \cite{Liu2022Reference-freeRatio}. These metrics measure the quality of summarizations with respect to the original document, which is the groundtruth descriptions of each test video, and they also take the summarization length into consideration. The metrics are calculated over the test set to obtain the mean summarization quality. We compare ChatGPT summarizations from our prediction method (top-11 sum.), randomly chosen sentences from the training set (random sum.), and ChatGPT summarizations from the groundtruth descriptions (gt. sum.), and they are compared under different numbers of summary sentences. The ranking of these summarizations are consistent across metrics and the result is not surprising, with groundtruth summarization being the most accurate, followed by the top-11 summarization, and they both clearly surpass random summarization.
However, we need to note that current large language models like ChatGPT may still have hallucination problems even though retrieved reference descriptions have been provided. Given the fast evolution of large models, we believe the employment of them in the DREAM task will become more significant and fruitful.
\subsection{Cross-dataset Evaluation}
We employ two external datasets to test the generalization ability of some leading methods from our benchmark. The FaceForensics++ (FF++) dataset \cite{Rossler2019Faceforensics++:Images} is composed of data from four deepfake and face reenactment methods and their corresponding real videos. We sample 100 videos from the C23 quality subset of FF++, and each video is annotated by 10 annotators in-house to obtain their MOS values. The second dataset is from the Q-Eval-100K dataset \cite{Zhang2025Q-Eval-100K:Content} which includes data from 22 text-to-video generation models. This dataset poses a much larger challenge with its very different generation methods, e.g. Diffusion. We sample 207 videos from its public training set that includes valid human faces and use its original visual quality annotation as the target MOS. Note its annotation may be subject to relatively high variance, since each video is annotated by at least three annotators, meaning a portion of it has only three annotations. The generalization performance of representative methods is shown in Table \ref{tab_Outer}. For these cross-dataset results, we use all the DREAM dataset for training and repeat with 10 random initializations for deep methods or 10 sets of selected features for DFGC1st. 
\begin{table} [th] 
\centering
\caption{Cross-dataset evaluation on FF++ and Q-Eval-100K. }  \label{tab_Outer}
\renewcommand{\arraystretch}{1.0}
\renewcommand{\arraystretch}{1.3}
\scalebox{0.9}{%
\begin{tabular}{|c|c|c|c|c|c|c|}
\hline
\multirow{2}{*}{Method}    & \multicolumn{3}{c|}{FF++}   & \multicolumn{3}{c|}{Q-Eval-100K}   \\
\cline{2-7}
  &PLCC $\uparrow$ &SRCC $\uparrow$ &avr $\uparrow$ &PLCC $\uparrow$ &SRCC $\uparrow$ &avr $\uparrow$  \\
\hline
DFGC1st  &\makecell{\underline{0.580}\\ $\pm$ 0.044}   &\makecell{\underline{0.556}\\ $\pm$ 0.047}    &\makecell{\underline{0.568}\\ $\pm$ 0.045}     &\makecell{\underline{0.238}\\ $\pm$ 0.033}    &\makecell{\underline{0.218}\\ $\pm$ 0.038}    &\makecell{\underline{0.228}\\ $\pm$ 0.035} \\
 \hline
UNILJ  &\makecell{0.178\\ $\pm$ 0.073}   &\makecell{0.175\\ $\pm$ 0.076}    &\makecell{0.176\\ $\pm$ 0.074}     &\makecell{-0.032\\ $\pm$ 0.074}    &\makecell{-0.042\\ $\pm$ 0.072}    &\makecell{-0.037\\ $\pm$ 0.073} \\
 \hline
OPDAI   &\makecell{0.308\\ $\pm$ 0.036}     &\makecell{0.303\\ $\pm$ 0.042}     &\makecell{0.305\\ $\pm$ 0.039}     & \makecell{0.051\\ $\pm$ 0.059}    &\makecell{0.018\\ $\pm$ 0.064}    &\makecell{0.035\\ $\pm$ 0.061} \\
 \hline
Fast-VQA   &\makecell{0.349\\ $\pm$ 0.036}     &\makecell{0.323\\ $\pm$ 0.040}     &\makecell{0.336\\ $\pm$ 0.038}     & \textit{\makecell{0.199\\ $\pm$ 0.026}}    &\textit{\makecell{0.202\\ $\pm$ 0.027}}    &\textit{\makecell{0.200\\ $\pm$ 0.025}} \\
 \hline
DeQA-i  &\textbf{\makecell{0.678\\ $\pm$ 0.024}}     &\textbf{\makecell{0.685\\ $\pm$ 0.029}}      &\textbf{\makecell{0.681\\ $\pm$ 0.026}}     &\textbf{\makecell{0.354\\ $\pm$ 0.022}}     &\textbf{\makecell{0.339\\ $\pm$ 0.019}}     &\textbf{\makecell{0.347\\ $\pm$ 0.020}}  \\
\hline
DA-CLIP    &\textit{\makecell{0.373\\ $\pm$ 0.018}}  &\textit{\makecell{0.383\\ $\pm$ 0.021}}     &\textit{\makecell{0.378\\ $\pm$ 0.019}}     &\makecell{0.141\\ $\pm$ 0.055}     &\makecell{0.113\\ $\pm$ 0.061}     &\makecell{0.127\\ $\pm$ 0.057}   \\
\hline
\end{tabular} 
} 
\end{table}

It is clear that most methods have trouble generalizing to the quality assessment of cross-domain datasets, and AI-generated videos are more different and harder to generalize compared to new deepfake data. The performance rank of different methods is also different from that of in-domain test in Table \ref{tab_Results}, with the best and second methods being DeQA-i and DFGC1st respectively, followed by the proposed DA-CLIP and Fast-VQA. Although the finetuning based methods, i.e., OPDAI and UNILJ, perform second and third in-domain, they tend to be more over-fitted. We conjecture that DeQA-i benefits from its VLM pretraining on large-scale and diverse datasets that also  include IQA/VQA-related ones \cite{Chen2024ExpandingScaling}, while DFGC1st relies on fixed pretrained features with a simple regression head, making them less affected by dataset-specific biases in finetuning. These observations indicate some trade-off between in-domain and cross-domain performances, and we think that the generalization of DREAM requires more future efforts in aspects of constructing diverse datasets and designing dedicated methods.

\section{Limitation}
There are two limitations that following work should be aware of. The first one is annotator demographic homogeneity. Although we recruited  3,500 annotators, which is a large number, they are all from China. Our annotation campaign is a practical design under budget and resource limitations, but this could lead to potential annotation bias from homogeneous backgrounds of familiarity and culture. We make our dataset available to the research community, so that it is possible for following researchers to conduct independent annotations and cross validate. Second, we only focused on visual deepfake assessment, but there are also audio deepfake and cross-modal audiovisual deepfake, whose realism assessment is also important and demands more future attention.

\section{Conclusion}
In this paper, we focus on a new task of deepfake photorealism assessment, and we propose a comprehensive benchmark called DREAM, that is comprised of a deepfake video dataset of diverse quality, a large scale annotation 
and a comprehensive evaluation and analysis of 18 representative photorealism assessment methods, including recent large vision language model based methods and a newly proposed description-aligned CLIP method. Through the experiments, we can see that reasonable accuracy for in-domain MOS regression can be achieved, but cross-dataset generalization needs to be further improved.
Benefiting from the textual descriptions, the cross-modal alignment of visual and textual cues can be better learned, which we think is a very promising direction that deserve more future investigation. Finally, we believe the DREAM benchmark and insights included in this study can lay the foundation for future research in this direction and other related areas.



 
\bibliographystyle{IEEEtran}
\bibliography{IEEEabrv, ./references}

\vfill

\end{document}